\documentclass[11pt,a4paper]{article}

\usepackage[utf8]{inputenc}
\usepackage[margin=1in]{geometry}
\usepackage{amsmath, amsthm, amssymb}
\usepackage{natbib} 
\usepackage{hyperref}
\usepackage{algorithm2e} 
\usepackage{bbm} 

\newtheorem{theorem}{Theorem}[section]
\newtheorem{lemma}[theorem]{Lemma}
\newtheorem{proposition}[theorem]{Proposition}
\newtheorem{definition}[theorem]{Definition}
\newtheorem{example}[theorem]{Example}
\newtheorem{corollary}[theorem]{Corollary}

\newcommand{\EM}{\ensuremath{\mathcal{EM}}}
\newcommand{\Lt}{\ensuremath{\Lambda}}
\newcommand{\AL}{\ensuremath{\mathrm{AL}}}
\newcommand{\PD}{\ensuremath{\mathrm{PD}}}
\newcommand{\VC}{\ensuremath{\mathrm{VC}}}
\newcommand{\PC}{\ensuremath{\mathcal{C}}}
\newcommand{\NR}{\ensuremath{E}}
\newcommand{\Lf}{\ensuremath{V_f}}
\newcommand{\NL}{\ensuremath{V_p}}
\newcommand{\DE}{\ensuremath{E_0}}
\newcommand{\Pa}{\ensuremath{P}}
\newcommand{\ReSet}{\ensuremath{R}}
\DeclareMathOperator{\Ra}{rank}
\DeclareMathOperator{\Dp}{dep}
\DeclareMathOperator{\Ar}{ar}
\newcommand{\Li}{\ensuremath{\mathrm{L}}}
\newcommand{\CL}{\ensuremath{\mathrm{L}_{\mathfrak{P}}}}
\newcommand{\Tree}{\ensuremath{=(V,v_0,\Pa,\mathbf{x},\mathbf{y},\mathbf{h},\DE)}}
\newcommand{\YP}{\ensuremath{Y^{\mathfrak{P}}}}

\newcommand{\apm}{\ensuremath{\mathcal{PM}}}

\newcommand{\cc}{\ensuremath{\mathcal{C}}}

\newcommand{\hh}{\ensuremath{\mathcal{H}}}

\newcommand{\mm}{\ensuremath{\mathcal{M}}}

\DeclareMathOperator*{\argmin}{arg\,min}
\DeclareMathOperator*{\argmax}{arg\,max}

\let\Pr\relax
\DeclareMathOperator{\Pr}{Pr}

\title{Ambiguous Online Learning}

\author{
  Vanessa Kosoy\thanks{Faculty of Mathematics, Technion – Israel Institute of Technology, Haifa, Israel. Email: \texttt{vanessa@alter.org.il}}
}
\date{\today}

\begin{document}

\maketitle
\begin{abstract}
We propose a new variant of online learning that we call ``ambiguous online learning". In this setting, the learner is allowed to produce multiple predicted labels. Such an ``ambiguous prediction" is considered correct when at least one of the labels is correct, and none of the labels are ``predictably wrong". The definition of ``predictably wrong" comes from a hypothesis class in which hypotheses are also multi-valued. Thus, a prediction is ``predictably wrong" if it's not allowed by the (unknown) true hypothesis. In particular, this setting is natural in the context of multivalued dynamical systems, recommendation algorithms and lossless compression. It is also strongly related to so-called ``apple tasting". We show that in this setting, there is a trichotomy of mistake bounds: up to logarithmic factors, any hypothesis class has an optimal mistake bound of either $\Theta(1)$, $\tilde{\Theta}(\sqrt{N})$ or $\Theta(N)$.
\end{abstract}


\section{Introduction}
In classical online learning theory, every hypothesis is a mapping $h:X \to Y$ from the domain $X$ to the label set $Y$. This assumes that, in some sense, every instance $x \in X$ has an unambiguous ``correct" label $h^*(x) \in Y$ associated with it. However, this is not always a safe assumption:

\begin{example}
\label{ex_car}
We want to predict the motion of a car, as constrained by the laws of physics and the car's design. However, we don't want to trust any assumption about the behavior of the driver. If $Y$ is the configuration space of the car (e.g. center-of-mass coordinates, angular orientation and steering wheel angle), coarse-grained to our measurement accuracy, and $X:=Y^*$ is the set of possible car motion histories, the physics we wish to learn can be described as a mapping $h^*:X \to 2^Y$. For example, if $x$ describes a car at rest, then $h^*(x)$ contains a label corresponding to the car staying at rest, but also labels corresponding to the driver pushing the gas pedal. However, even if the driver floors the gas pedal, the car can only accelerate that much within the relevant time interval, which constrains the labels that fall inside $h^*(x)$.
\end{example}

More generally, such multivalued hypotheses can be used to describe any \emph{multivalued} dynamical system\footnote{Meaning a system under some unpredictable external influence: see e.g. \citep{brogliatoT2020}}.

Another example is systems which offer multiple choices to the user, but only some choices are appropriate in any given context:

\begin{example}
The user is writing text in a messaging app and occasionally inserts emojis. The algorithm learns to offer the set of emojis potentially appropriate in a given context. Here, $X$ is the record of the previous conversation, $Y$ is the set of all emojis and $h^*:X \to 2^Y$ assigns to each conversation the set of emojis that might plausibly follow it\footnote{In real life, there might be rare occasions on which the user wants to use an emoji which doesn't normally make any sense in the given context. This can be modeled by e.g. adding some degree of ``noise". However, in the present work we ignore such complications to focus on the mathematically simpler case of deterministic hypotheses.}.
\end{example}

More examples come from data following some unknown formal syntax:

\begin{example}
We need to losslessly compress a stream of data that is likely to follow a particular syntax (e.g., it is a Python source file), but we don't know which. If $Y$ is the alphabet and $X:=Y^*$ represents the data seen so far, the unknown syntax can be represented as a mapping $h^*:X \to 2^Y$, where $h^*(x)$ is the set of symbols that can legally follow the string $x$ according to the syntax. For example, if a Python file starts with ``2=", the next symbol has to be '='. On the other hand, ``2==" can be followed by multiple symbols (e.g., '3' or 'x', but not '\#').
\end{example}

In all such examples, we could alternatively try to model the uncertainty about the label in terms of probabilities. That is, postulate a mapping $h^*:X \to \Delta Y$ instead of $h^*:X \to 2^Y$. If the underlying process can be regarded as IID samples from a fixed distribution (or at least as a stationary stochastic process), these probabilities have a natural frequentist interpretation. However, if the underlying process is non-stationary (s.t. online learning is called for), the meaning of these probabilities is not always clear\footnote{In principle, we can apply some Bayesian notion of probability, but it raises difficult questions about the choice of prior. For further discussion of the interpretation of probability and alternative representations of uncertainty, see e.g. \citep{hajekH2016}.}. Moreover, even if the probability distributions are defined in principle, they might be much harder to compute or learn than their supports. More generally, we could combine probabilistic uncertainty with the ``ambiguity" we study here by allowing sets of probability distributions (as in e.g. \cite{kosoy2024imprecisemultiarmedbandits}). However, for the sake of simplicity we consider only sets of deterministic labels in the present work. With this motivation, we consider an online learning setting with a class of multivalued hypotheses $\mathcal{H} \subseteq \{X \to 2^Y\}$. The learning algorithm is also allowed to output sets of labels. Given true hypothesis $h^* \in \mathcal{H}$, instance $x \in X$, the algorithm's prediction $\alpha \subseteq Y$ and the actual observed label $y \in Y$, the algorithm's success criterion is
\[ y \in \alpha \subseteq h^*(x) \]
The condition $y \in \alpha$ requires that the prediction is sound, whereas the condition $\alpha \subseteq h^*(x)$ requires that the prediction is ``complete", i.e. no weaker than what knowledge of $h^*$ would allow\footnote{If we interpret a prediction $\alpha$ as making the logical statement $y\in\alpha$ about the following label $y$, then a prediction is sound iff the statement is \emph{true} and complete iff the statement is as \emph{strong} as can be reasonably expected.}. Any prediction for which the success criterion doesn't hold is considered to be a ``mistake". In classical online learning theory, the maximin number of mistakes for any hypothesis class, assuming realizability, is given by the Littlestone dimension. In particular, the asymptotic as a function of the time horizon $N$ can be either $O(1)$ (when the Littlestone dimension is finite) or $\Omega(N)$ (when the Littlestone dimension is infinite). On the other hand, for ambiguous online learning, we prove a trichotomy of possible mistake bounds: depending on the hypothesis class, the number of mistakes can be $O(1)$, $\tilde{\Theta}(\sqrt{N})$ or $\Omega(N)$. In the $O(1)$ case, the exact number of mistakes is given by a combinatorial invariant that we call ``ambiguous Littlestone dimension". In the $\tilde{\Theta}(\sqrt{N})$ case, the coefficient can be bounded using the ordinary Littlestone dimension for the label set $2^Y$, and another combinatorial ``dimension" associated with the set of subsets of $Y$ that hypotheses can produce.

\subsection*{Related Work}
\cite{moranSTY23} study a version of online learning where the learner is allowed to produce multivalued predictions and hypotheses can also be multivalued\footnote{For hypotheses, they consider the even more general notion of ``pattern classes".}, similarly to our setting. However, their success criterion is weaker: given a prediction $\alpha \subseteq Y$ and an observed label $y \in Y$, a correct prediction only requires that $y \in \alpha$. To make this non-trivial, they require that $|\alpha| \le k$ for some fixed $k \in \mathbb{N}$, whereas we have no such constraint. 

\cite{cohenMMS24} consider offline (PAC) learning of graphs with a loss function derived from \emph{precision and recall}. In particular, their setting can describe a multivalued hypothesis class (in this case the graph is bipartite, with $X$ vertices and $Y$ vertices), and the learner's prediction is also multivalued (a graph). Their success criterion is effectively stronger than our own: in our notation, their loss only vanishes when $\alpha=h^*(x)$. However, they assume that $X$-instances are IID samples from some fixed distribution and, even more importantly, $Y$-labels are uniform IID samples from $h^*(x)$. The latter assumption is critical: in our setting we cannot guarantee convergence to $\alpha=h^*(x)$ because the adversary can choose to never select certain elements of $h^*(x)$.

Another setting with multivalued prediction is \emph{conformal} prediction (see e.g. \cite{fontana2023}). In conformal prediction, the goal is calibrating the size of the prediction sets to achieve a given error rate. This is very different from our setting, where the error rate is required to vanish asymptotically. Moreover, conformal prediction requires an IID assumption. (On the other hand, conformal prediction doesn't assume realizability and can use any probabilistic prediction algorithm as a baseline.)

Apple tasting (see \cite{hemboldLL2000}, \cite{ramanSRT24}, \cite{chaseM24}) is a variant of online learning with labels $Y=\{0,1\}$ where the learner only receives feedback when predicting $1$. This setting is known to have a trichotomy of mistake bounds similar to our own. In Section~\ref{sec_apple} we show that there is a formal relationship between the two frameworks. Specifically, there is a way to translate an apple tasting hypothesis class into an ambiguous online learning hypothesis class in a way that preserves the minimax mistake bound, explaining why the trichotomies are the same. 

\cite{alonHHM2021} introduced a theory of online learning for \emph{partial} concept classes. In this setting, a hypothesis $h \in \hh$ is allowed to be undefined on some instances $x \in X$, meaning that if the hypothesis is true, these instances cannot appear. This is a simple special case of our setting, where ``$h$ is undefined on $x$" corresponds to $h(x)=\emptyset$. In fact, we use their generalization of Littlestone dimension to state the $\tilde{O}(\sqrt{N})$ upper bound (Theorem~\ref{thm_upper_sqrt}). 

A recent line of research studies multiarmed bandits and reinforcement learning with multivalued hypotheses (see \cite{kosoy2024imprecisemultiarmedbandits}, \cite{appelK25}). That's much more general in the sense that their hypotheses produce sets of \emph{probability distributions}\footnote{Even though it's also less general because the bandits studied there are not contextual, so there's no analogue of $X$.}. However, the success metric there is very different: it involves the regret of observable attained reward relatively to the maximin reward of the true hypothesis. 

The structure of the rest of the paper is as follows. In Section~\ref{sec_setting} we formally define the setting and the mistake bounds we are interested in. In Section~\ref{sec_inv} we define various combinatorial invariants that appear in our bounds are explore their basic properties. In Section~\ref{sec_main} we state our main results about mistake bounds in ambiguous online learning. In Section~\ref{sec_aoa} we describe the algorithm (AOA) that achieves minimax mistakes (and in particular certifies the upper bound in the $O(1)$ case). In Section~\ref{sec_waa} we describe the algorithm (WAA) that certifies the upper bound in the $\tilde{O}(\sqrt{N})$ case. Finally, in Section~\ref{sec_apple} we discuss the relation to apple tasting, and additional lower bounds related to that.

\section{Setting}
\label{sec_setting}
\label{sec:setting}
Fix a set $X$. We call $X$ the ``domain" and its elements ``instances". Fix another set $Y$, which we assume to be finite\footnote{The infinite case can also be studied, but this introduces additional complications and we won't attempt it here.}. We call its elements ``labels". Finally, consider a non-empty set $\mathcal{H} \subseteq \{X \to 2^Y\}$ whose elements are ``hypotheses". We require that for every $h \in \hh$, there exists $x \in X$ s.t. $h(x) \neq \emptyset$.

We are interested in the following game between the ``learner" and the ``adversary". The adversary chooses some $h^*\in\mathcal{H}$ in the beginning of the game, which is unknown to the learner. Each round proceeds as follows:

\begin{enumerate}
    \item The adversary chooses $x \in X$ and shows it to the learner. 
    \item The learner makes a prediction $\alpha\in2^Y$. 
    \item The adversary chooses a label $y\in h^*(x)$ which is revealed to the learner. 
    \item The learner is considered to have made a mistake when either $y\not\in\alpha$ or $\alpha\setminus h^*(x)\ne\emptyset$.
\end{enumerate}

Crucially, mistakes are \emph{not} part of the feedback. The learner doesn't know whether it made a mistake: in the case $y\not\in\alpha$, the learner can infer it, but in the case $\alpha\setminus h^*(x)\ne\emptyset$ it cannot, since it doesn't know $h^*$.

Formally, a (deterministic) \emph{learner} is a mapping $A:(X \times Y)^* \times X \to 2^Y$. Here, the notation $\Sigma^*$ means words over the alphabet $\Sigma$. In this case, the first argument of $A$ stands for the past observed history of instances and labels, and the second argument for the current instance for which we need to predict the label.

Given a learner $A$, a hypothesis $h:X \to 2^Y$, a number\footnote{We use the convention $0 \in \mathbb{N}$.} $n \in \mathbb{N}$ and a trace $xy \in (X \times Y)^n$, we define the set of \emph{mistakes} that $A$ makes on $xy$ relative to $h$ by
\[ \mathcal{M}^A_h (xy) := \{k < n \mid y_k \notin A(xy_{:k},x_k) \text{ or } A(xy_{:k},x_k) \not\subseteq h(x_k)\} \]
Here, $w_{:k}$ stands for the prefix of length $k$ of a word or sequence $w$, $w_k$ stands for the $k$-th symbol in the word (starting from 0, i.e. $w=w_0 w_1 \dots$), and $x$ and $y$ are the projections of $xy$ to $X^n$ and $Y^n$ respectively.

We call $y_k \notin A(xy_{:k},x_k)$ an \emph{overconfidence mistake} and $A(xy_{:k},x_k) \not\subseteq h(x_k)$ an \emph{underconfidence mistake}.

Given $h:X \to 2^Y$ and $xy \in (X \times Y)^n$, $xy$ is said to be \emph{compatible} with $h$ when for all $k<n$, $y_k \in h(x_k)$. We denote by $\mathcal{C}_h (n)$ the set of all such $xy$. We will also use the notations
\begin{align*}
\mathcal{C}_{\mathcal{H}}(N) &:= \bigcup_{h \in \mathcal{H}} \mathcal{C}_h (N) \\
\mathcal{M}^A_h (N) &:= \max_{xy \in \mathcal{C}_h (N)} |\mathcal{M}^A_h (xy)| \\
\mathcal{M}^A_{\mathcal{H}} (N) &:= \max_{h \in \mathcal{H}} \mathcal{M}^A_h (N) \\
\mathcal{M}^*_{\mathcal{H}} (N) &:= \min_A \mathcal{M}^A_\mathcal{H} (N)
\end{align*}
Our main goal will be determining the possible asymptotics of $\mathcal{M}^*_{\mathcal{H}} (N)$ (the minimax mistake bound) and understanding how it depends on $\mathcal{H}$. Notice that the definition of $\mathcal{M}^*_{\mathcal{H}} (N)$ implies realizability (the mistakes are defined w.r.t. a ``true hypothesis" s.t. the observed sequence is compatible with it). We will not consider the nonrealizable case in this work.

A possible variant is counting underconfidence mistakes and overconfidence mistakes with different weights. However, this at most changes the bound by a constant factor, and in particular doesn't affect the asymptotic classification. For the sake of simplicity, we will treat both kinds of mistakes symmetrically.

It's also possible to consider a \emph{randomized} learner $R:(X \times Y)^* \times X \to \Delta 2^Y$. Here, $\Delta \Sigma$ is our notation for the set of probability distribution over a set $\Sigma$. In this case, we can analogously define the \emph{expected} number of mistakes. Given a randomized learner $R$, a hypothesis $h:X \to 2^Y$, some $n \in \mathbb{N}$ and a trace $xy \in (X \times Y)^n$, we define the expected number of mistakes that $R$ makes on $xy$ relative to $h$ to be\footnote{Instead of static traces $xy$, it is possible to consider a setting in which the adversary can dynamically choose instances depending on the learner's predictions. Such a setting can only be harder for a randomized learner, but is equivalent for a deterministic learner. Therefore, the maximin mistake bound in that setting would lie between the lower bound of Proposition~\ref{prp_random} and the upper bound of Theorem~\ref{thm_O1}.}
\[ \EM^R_h (xy) := \sum_{k<n} \Pr_{\alpha \sim R(xy_{:k},x_k)}[y_k \notin \alpha \text{ or } \alpha \not\subseteq h(x_k)] \]
We also define $\EM_h^R (N)$, $\EM_{\hh}^R (N)$ and $\EM_{\hh}^* (N)$ analogously to the deterministic case. Obviously, $\EM^*_{\hh} (N) \le \mathcal{M}^*_{\hh} (N)$, but this inequality might be strict. In this work, we mostly focus on deterministic learners, but also demonstrate a lower bound on $\EM^*_{\hh} (N)$ which limits how much it can improve on $\mathcal{M}^*_{\hh} (N)$ to a factor that is constant in $N$.

\section{Combinatorial Invariants}
\label{sec_inv}
\label{sec:invariants}
In order to state our main theorem, we will need several combinatorial invariants. We start with the definition of ``ambiguous Littlestone dimension". Analogously to classical Littlestone dimension, the definition comes from a notion of ``shattered tree". However, these trees need not be perfect or binary.


We will need some notation for trees.

\begin{definition}
A \emph{rooted tree} is a tuple $T:=(V,v_0,\Pa)$, where $V$ is a finite set (called \emph{vertices}), $v_0 \in V$ is called the \emph{root}, and $\Pa:V \setminus \{v_0\} \to V$ is s.t. for any $v \in V$, there exists $k \in \mathbb{N}$ with $\Pa^k(v)=v_0$. This\footnote{It is easy to see that $k$ is unique for any given $v$.} $k$ is called the \emph{depth} of $v$. $\Pa(v)$ is called the \emph{parent} of $v$.
\end{definition}

We will denote by $\NR := V \setminus \{v_0\}$ the set of non-root vertices\footnote{Equivalently, we can think of $\NR$ as the \emph{edges} of the tree, with each edge corresponding to its child vertex.}, by $\Lf := V \setminus \Pa(\NR)$ the set of leaves and by $\NL := V \setminus \Lf$ the set of non-leaf vertices. We will also denote by $\Pa^*: V \to 2^V$ the mapping that assigns each vertex its set of ascendants including itself, i.e. $\Pa^*(v):=\{v,\Pa(v),\Pa^2(v)\dots\}$. 

The \emph{depth} of $T$ is defined to be 

$$ \Dp(T):=\max_{u \in \Lf} |\Pa^*(u)| - 1 $$

The \emph{arity} of $T$ is defined to be
\[ \Ar(T):=\max_{v \in \NL} |\Pa^{-1}(v)| \]

\begin{definition}
An \emph{ambiguous shattered $\mathcal{H}$-tree} is a tuple $T=(V,v_0,\Pa,\mathbf{x},\mathbf{y},\mathbf{h},\DE)$, where $(V,v_0,\Pa)$ is a rooted tree, $\mathbf{x}:\NL \to X$, $\mathbf{y}:\NR \to Y$, $\mathbf{h}:\Lf \to \mathcal{H}$ are labelings for which we will use subscript notation (e.g. write $\mathbf{x}_v$ instead of $\mathbf{x}(v)$) and $\DE \subseteq \NR$. These need to satisfy the following conditions:
\begin{itemize}
    \item For any $v \in \NL$, $|\Pa^{-1}(v) \cap \DE| = 1$.
    \item For any $a,b \in \NR$, if $a \neq b$ but $\Pa(a)=\Pa(b)$ then $\mathbf{y}_a \neq \mathbf{y}_b$.
    \item For any $u \in \Lf$ and $a \in \Pa^*(u) \setminus \{v_0\}$: $\mathbf{y}_a \in \mathbf{h}_u (\mathbf{x}_{\Pa(a)})$.
\end{itemize}
Consider $u \in \Lf$ and $a \in \Pa^*(u) \setminus \{v_0\}$, and denote $v:=\Pa(a)$. The edge $a$ is said to be \emph{relevant} to $u$ when either $a \notin \DE$ or there exists $b \in \Pa^{-1}(v)$ s.t. $\mathbf{y}_b \notin \mathbf{h}_u (\mathbf{x}_v)$. We denote the set of such $a$ by $\ReSet_T(u)$. The \emph{rank} of $T$ is defined to be
\[ \Ra(T):=\min_{u \in \Lf}|\ReSet_T (u)| \]
\end{definition}

As we noted above, ambiguous shattered trees need not be perfect, i.e. different leaves can have different depth. On the other hand, it's always possible to assume that all leaves have the same rank.

Conceptually, an ambiguous shattered tree $T$ represents a possible strategy for the adversary to force at least $\Ra(T)$ mistakes. This strategy proceeds by traversing the tree from the root to a leaf. The vertex labels $\mathbf{x}$ represent the instances the adversary would choose, the edge labels $\mathbf{y}$ represent the labels the adversary would choose, and the final leaf label $\mathbf{h}_u$ represents the hypothesis w.r.t. which the resulting trace is realizable and the mistakes are counted. The edge set $\DE$ tells the adversary in which direction to go down the tree when the learner predicts a set that contains all the edge labels attached to the current vertex. When an edge $a$ is ``relevant" to a leaf $u$, this implies that the learner accrues a mistake when the process goes through $a$ and ends on $u$.

\begin{proposition}
\label{prp_uniform_rank}
For any ambiguous shattered $\mathcal{H}$-tree $T=(V,v_0,\Pa,\mathbf{x},\mathbf{y},\mathbf{h},\DE)$, there exists another ambiguous shattered $\mathcal{H}$-tree of the form $T'=(V' \subseteq V,v_0,\Pa|_{V'}, \mathbf{x}|_{\NL'}, \mathbf{y}|_{\NR'}, \mathbf{h}', \DE \cap \NR')$ s.t. for every $u \in \Lf'$, $|\ReSet_{T'}(u)|=\Ra(T)$.
\end{proposition}

The proofs of the propositions in this section are given in Appendix~\ref{apx_com_props}. The \emph{ambiguous Littlestone dimension} of $\mathcal{H}$ (denoted $\AL(\mathcal{H})$) is defined to be the maximal rank of an ambiguous shattered $\mathcal{H}$-tree, or $\infty$ if the rank is unbounded. We also denote by $\AL(\mathcal{H},n)$ the maximal rank of an ambiguous shattered $\mathcal{H}$-tree of depth at most $n$. Obviously, $\AL(\mathcal{H}) = \lim_{n\to\infty} \AL(\mathcal{H},n)$.

For finite $\mathcal{H}$, we have the following upper bound:

\begin{proposition}
\label{prp_al_card}
$\AL(\mathcal{H}) \le |\mathcal{H}|-1$
\end{proposition}

Notice that, as opposed to classical Littlestone dimension, $\AL(\mathcal{H})$ is \emph{not} bounded by $O(\log |\mathcal{H}|)$. Indeed, we have:

\begin{example}
\label{ex_fin_deltas}
For any $n \ge 1$, let $X_n:=[n]$ and $Y=\{0,1\}$. For any $k < n$, define $h_{k,n}:[n] \to 2^{\{0,1\}}$ by
\[ h_{k,n} (x)= \begin{cases} \{1\} & \text{if } x=k, \\ \{0,1\} & \text{if } x \neq k \end{cases} \]
Let $\mathcal{H}_n:=\{h_{k,n} \mid k<n\}$. Then, $|\mathcal{H}_n|=n$ and $\AL(\mathcal{H}_n,n^2)=n-1$.
\end{example}

Here and elsewhere, we use the notation $[n]:=\{k \in \mathbb{N} \mid k < n\}$.

To verify Example~\ref{ex_fin_deltas}, we calculate the ambiguous Littlestone dimension by explicitly constructing an ambiguous shattered tree. This tree represents an adversary strategy which progresses through instances $x \in [n]$ starting from 0 and going up. When the learner predicts $\{1\}$, the adversary responds by showing the label 0 and moving to the next instance. In this case, the learner incurs an overconfidence mistake. When the learner predicts $\{0,1\}$, the adversary responds by showing the label 1 and choosing the same instance $k$ again. This incurs an underconfidence mistake against hypotheses $h_j$ with $j>k$.

\begin{proof}[of Example~\ref{ex_fin_deltas}]
For any $v \in \{0,1\}^*$, let $r_0(v)$ be the number of 0s in $v$ and $r_1(v)$ be the maximal $k$ s.t. $1^k$ is a suffix of $v$. Denote $r(v) := r_0(v)+r_1(v)$. Fix $n \ge 1$. Define $V$ by
\[ V := \{v \in \{0,1\}^* \mid r(v)<n \text{ and } \forall k<|v|: r(v_{:k}) < n-1\} \]
Let $v_0 := \lambda$. Let $\Pa(vi) := v$ for $i \in \{0,1\}$. Then, $T=(V,v_0,\Pa)$ is a rooted binary tree. Notice that
\[ \Lf = \{v \in \{0,1\}^* \mid r(v)=n-1 \text{ and } \forall k<|v|: r(v_{:k}) < n-1\} \]
In any $v \in V$ there are at most $n-1$ 0s, and between the 0s we have stretches of $1$s of length at most $n-1$. Hence, $|v| \le n-1+n(n-1) = n^2-1$ and $\Dp(T) \le n^2-1$.

Let $\mathbf{x}_v := r_0(v)$. Let $\mathbf{y}_{vi} := i$ for $i \in \{0,1\}$. Let $\mathbf{h}_u := h_{r_0(u),n}$. Finally, let $\DE := \{a \in \NR \mid \mathbf{y}_a = 1\}$. Then, $T^* \Tree$ is an ambiguous shattered $\hh_n$-tree.

For any $u \in \Lf$, we have
\[ \ReSet_T(u) = \{v0 \mid \exists w \in \{0,1\}^*: u=v0w\} \cup \{v1 \mid \exists k \ge 1: u=v1^k\} \]
Hence, $|\ReSet_T(u)| = r(u) = n-1$ and $\Ra(T^*) = n-1$, implying that $\AL(\hh_n,n^2) \ge n-1$. By Proposition~\ref{prp_al_card}, we also have $\AL(\hh_n,n^2) \le n-1$.
\end{proof}

To give another example, we show that flipping the hypothesis values in Example~$\ref{ex_fin_deltas}$ changes the dimension dramatically.

\begin{example}
\label{ex_small_al}
For any $n \ge 1$, let $X_n:=[n]$ and $Y=\{0,1\}$. For any $k < n$, define $h_{k,n}:[n] \to 2^{\{0,1\}}$ by
\[ h_{k,n} (x)= \begin{cases} \{0,1\} & \text{if } x=k, \\ \{1\} & \text{if } x \neq k \end{cases} \]
Let $\mathcal{H}_n:=\{h_{k,n} \mid k<n\}$. Then, $\AL(\mathcal{H}_n)=1$.
\end{example}

\begin{proof}
In order to see that $\AL(\mathcal{H}_n)\geq1$, consider the following ambiguous tree $T\Tree$. We take $V:=\{v_0,u_0,u_1\}$, $\Pa(u_0)=\Pa(u_1)=v_0$, $\mathbf{x}_{v_0}:=0$, $\mathbf{y}_{u_0}:=0$, $\mathbf{y}_{u_1}:=1$, $\mathbf{h}_{u_0}:=h_{0,n}$, $\mathbf{h}_{u_1}:=h_{1,n}$, $\DE:=\{u_1\}$. It is easy to see that $T$ is shattered and $\Ra(T)=1$.

Now, let's show that $\AL(\mathcal{H}_n)\leq1$. Consider \emph{any} ambiguous shattered tree $T\Tree$. We need to show that $\Ra(T)\leq 1$. If $\Ra(T)>0$ then there is a unique vertex $v_1 \in V$ s.t. $|\Pa^{-1}(v_1)|>1$ and for all $v\in \Pa^*(v_1)\setminus\{v_1\}$, $|\Pa^{-1}(v)|=1$. (The first vertex of degree 3 or more as we go down from the root; if there is no such vertex, all edges are irrelevant.) We must have $\Pa^{-1}(v_1)=\{a_0,a_1\}$ for some $a_0,a_1\in\NR$ s.t. $\mathbf{y}_{a_0}=0$ and $\mathbf{y}_{a_1}=1$. Consider the unique leaf $u\in\Lf$ s.t.  $a_0\in\Pa^*(u)$ and for any $b\in\Pa^*(u)$, if $a_0\in\Pa^*(b)\setminus\{b\}$ then $b\in\DE$. (That is, $u$ is the end of the $\DE$-path starting from $a_0$.) Let $x:=\mathbf{x}_{v_1}$ and $h:=\mathbf{h}_u$. We must have $0\in h(x)$ and therefore $h=h_{x,n}$. We will now show that $|R_T (u)|\leq1$ and hence $\Ra(T)\leq1$.

Consider any $b\in R_T (u)$. There are 3 possibilities:
\begin{itemize}
    \item $b=a_0$, in which case it might be that $b\in R_T (u)$.
    \item $b\in\Pa^* (v_1)$, in which case $|\Pa^{-1}(\Pa(b))|=1$ and hence $b\not\in R_T (u)$.
    \item $a_0\in\Pa^*(b)\setminus\{b\}$. Then, $b\in\DE$. Denote $v_2:=\Pa(b)$ and $x':=\mathbf{x}_{v_2}$. If $x'=x$ then $h(x')=\{0,1\}$ and hence $b\not\in R_T (u)$. On the other hand, if $x'\ne x$ then $h(x')=\{1\}$ and hence $b\in R_T (u)$ iff there exists $c\in\Pa^{-1} (v_2)$ s.t. $\mathbf{y}_c=0$. However, such $c$ cannot exist. Indeed, assume to the contrary that $c$ exists. Choose any $u'\in\Lf$ s.t. $c\in\Pa^*(u')$ and denote $h':=\mathbf{h}_{u'}$. Since $a_0\in\Pa^* (u')$, we must have $0\in h'(x)$ and therefore $h'=h_{x,n}$. But then $h'(x')=\{1\}$ which is inconsistent with $\mathbf{y}_c=0$. We conclude that $b\not\in R_T (u)$.
\end{itemize}

We see that $R_T (u)$ has at most one element ($a_0$).
\end{proof}

\subsection*{Pivot Dimension}

Our next invariant of interest is related to those subsets of $Y$ that appear in hypotheses.

\begin{definition}
We define a \emph{$Y$-lattice}\footnote{It is easy to see that any such $\Lt$ is indeed a lattice for the partial order $A \subseteq B$.} to be $\Lt \subseteq 2^Y$ s.t.
\begin{itemize}
    \item $Y \in \Lt$
    \item For any $A,B \in \Lt$, also $A \cap B \in \Lt$.
\end{itemize}
Given $A \subseteq Y$, define the \emph{$\Lt$-hull} of $A$ to be
\[ A^{(\Lt)}:=\bigcap_{B \in \Lt:A \subseteq B} B \]
Given $A \subseteq Y$, define the \emph{$\Lt$-complexity} of $A$ by
\[ \PC_{\Lt}(A):=\min_{B \subseteq A: A \subseteq B^{(\Lt)}} |B| \]
The \emph{pivot dimension} of $\Lt$ is defined to be\footnote{Note that this is not the same as $\PC_{\Lt}(Y)$. Because, $A \subseteq B$ doesn't imply $\PC_{\Lt}(A) \le \PC_{\Lt}(B)$ in general.}
\[ \PD(\Lt):=\max_{A \in \Lt} \PC_{\Lt}(A) \]

\end{definition}

In words, the $\Lt$-hull of $A$ is the minimal element of $\Lt$ containing $A$. The $\Lt$-complexity of $A$ is the minimal size of a subset $B$ of $A$ s.t. the $\Lt$-hull of $B$ contains all of $A$. The pivot dimension is the maximal $\Lt$-complexity of any set in $\Lt$.

Now, let $\Lt(\mathcal{H}) \subseteq 2^Y$ be the unique minimal $Y$-lattice s.t. for any $h \in \mathcal{H}$ and $x \in X$, $h(x) \in \Lt(\mathcal{H})$. Our invariant of interest is $\PD(\Lt(\mathcal{H}))$. Abusing notation, we will denote it just $\PD(\hh)$.

We can relate $\PD(\Lt)$ to $\VC(\Lt)$, the VC dimension of $\Lt$\footnote{For the definition of VC dimension, see e.g. \cite{shalevB2014}.}:

\begin{proposition}
\label{prp_upd_card}
$\PD(\Lt) \le \VC(\Lt)$
\end{proposition}

In particular, $\PD(\Lt) \le \log|\Lt| \le |Y|$. (Here and everywhere else, $\log$ stands for the logarithm to base 2.) These bounds are tight, as can be seen from the case $\Lambda=2^Y$ where $\PD(\Lt)=|Y|$. On the other hand, it's possible to have $\PD(\Lt) \ll \VC(\Lt) \ll \log |\Lt|$. For example:

\begin{example}
\label{ex_box}
Consider any integers $d \ge 1$ and $n \ge 3$. Let $Y_{d,n}:=[n]^d$. Define $\Lt_{d,n}$ by
\[ \Lt_{d,n}:=\{\{y \in Y \mid \forall i<d: a_i \le y_i \le b_i\} \mid a,b \in Y\} \]
Then, $\PD(\Lt_{d,n})=2$, $\VC(\Lt_{d,n})=2d$ and $\log |\Lt_{d,n}| \approx 2d \log n$ (the latter holds up to an additive constant).
\end{example}

Note that Example~\ref{ex_box} has a natural interpretation if we think of $Y$ as the coarse-grained $d$-dimensional state space of a dynamical system (similarly to Example~\ref{ex_car}) and we only try to predict each coordinate separately (rather than their mutual dependency). 

$\VC(\Lt)$ is in itself relevant to our setting, due to the following:

\begin{proposition}
\label{prp_arity_upd}
For any $n \in \mathbb{N}$ and $r \le \AL(\hh,n)$, there exists an ambiguous shattered $\hh$-tree $T$ s.t. $\Ra(T) \ge r$, $\Dp(T) \le n$ and $\Ar(T) \le \VC(\Lt(\hh))+1$.
\end{proposition}

It is also possible to bound $\AL(\mathcal{H})$ in terms of $|X|$ and lattice structure of $\Lt(\mathcal{H})$.

\begin{proposition}
\label{prp_al_dom}
Let $l$ be the length\footnote{The length of a lattice is the cardinality of its longest chain minus one. See e.g. \citep{gratzer2003}.} of the lattice $\Lt(\mathcal{H})$. Then, $\AL(\mathcal{H}) \le l |X|$.
\end{proposition}

In particular, $\AL(\mathcal{H}) \le |Y||X|$. This is again weaker than the classical bound $\Li(\mathcal{H}) \le |X|$.

The last invariant we need is the Littlestone dimension of $\mathcal{H}$, when we regard it a class of \emph{partial} functions from $X$ to $2^Y \setminus \{\emptyset\}$ in the obvious way. The Littlestone dimension of a class of partial functions (introduced in \cite{alonHHM2021}) is defined using shattered trees in the ordinary way, see Appendix~\ref{apx_partial} for details. We denote it $\CL(\mathcal{H})$.

For the usual reason, for finite $\mathcal{H}$, we have $\CL(\mathcal{H}) \le \log |\mathcal{H}|$. There is also a relation between classical shattered trees and ambiguous shattered trees.

\begin{proposition}
\label{prp_CL_vs_AL}
For any (classical) shattered $\mathcal{H}$-tree $T=(V,v_0,\Pa,\mathbf{x},\mathbf{y}:\NR \to 2^Y,\mathbf{h})$, there exists an ambiguous shattered $\mathcal{H}$-tree of the form $T'=(V,v_0,\Pa,\mathbf{x},\mathbf{y}':\NR \to Y,\mathbf{h},\DE)$ s.t. $\Ra(T')=\Dp(T')=\Dp(T)$.
\end{proposition}

As a trivial corollary, we infer that $\CL(\mathcal{H}) \le \AL(\mathcal{H})$.

\section{Main Results}
\label{sec_main}
\label{sec:results}
We are now ready to state our main results on mistake bounds in ambiguous online learning. First, we establish the $\Theta(1)$ branch of the trichotomy.

\begin{theorem}
\label{thm_O1}
For all $N \in \mathbb{N}$, $\mm^*_{\hh}(N) = \AL(\hh,N)$.
\end{theorem}

(See Appendix~\ref{apx_O1} for the proof.) In particular, $\lim_{N\to\infty} \mm^*_{\hh}(N) = \AL(\hh)$. If $\AL(\hh) < \infty$ then $\mm^*_{\hh} = \Theta(1)$.

Now, we consider the separation between the $\Theta(1)$ branch and the $\tilde{\Theta}(\sqrt{N})$ branch.

\begin{theorem}
\label{thm_compact}
For any $N \in \mathbb{N}$, if there exists an ambiguous shattered $\hh$-tree $T$ with $\Ra(T) \ge N^2$, then there exists an ambiguous shattered $\hh$-tree $T'$ with $\Ra(T') \ge N$ and $\Dp(T') \le N^2$.
\end{theorem}

(See Appendix~\ref{apx_compact} for the proof.) Theorem~\ref{thm_O1} and Theorem~\ref{thm_compact} immediately imply

\begin{corollary}
\label{crl_lower_sqrt}
If $\AL(\hh)=\infty$, then for all $N \in \mathbb{N}$, $\mm^*_{\hh}(N) \ge \lfloor\sqrt{N}\rfloor$.
\end{corollary}

To establish the $\tilde{\Theta}(\sqrt{N})$ branch, we also need an upper bound:

\begin{theorem}
\label{thm_upper_sqrt}
There exists some global constant $C>0$ s.t. for any $N \ge 1$,
\[ \mm^*_{\hh}(N) \le C \cdot \PD(\hh)\sqrt{\CL(\hh)N \log(|\Lt(\hh)|N)} \]
\end{theorem}

(See Appendix~\ref{apx_red} for the proof.) By Corollary~\ref{crl_lower_sqrt} and Theorem~\ref{thm_upper_sqrt}, whenever $\AL(\hh)=\infty$ and $\CL(\hh) < \infty$, we have $\mm^*_{\hh}(N) = \tilde{\Theta}(\sqrt{N})$. To see that this can indeed happen, consider the following:

\begin{example}
Let $X:=\mathbb{N}$, $Y:=\{0,1\}$ and $\mathcal{H}=\{h_k \mid k \in \mathbb{N}\}$, where $h_k:\mathbb{N} \to 2^{\{0,1\}}$ is defined by
\[ h_k(x) = \begin{cases} \{1\} & \text{if } x=k, \\ \{0,1\} & \text{if } x \neq k \end{cases} \]
Then, $\AL(\hh)=\infty$ and $\CL(\hh)=1$. To see that $\AL(\hh)=\infty$, we can use the ambiguous shattered trees from Example~\ref{ex_fin_deltas} for arbitrarily large $n$. To see that $\CL(\hh)=1$, notice that a classical online learner that memorizes past instances and predicts $\{0,1\}$ on new instances can only make 1 mistake.
\end{example}

We also need to separate the $\tilde{\Theta}(\sqrt{N})$ branch from the $\Theta(N)$ branch. As an easy corollary of Theorem~\ref{thm_O1} and Proposition~\ref{prp_CL_vs_AL}, we get

\begin{corollary}
If $\CL(\hh)=\infty$, then for all $N \in \mathbb{N}$, $\mm^*_{\hh}(N)=N$.
\end{corollary}

In classical theory there are plenty of examples with $\CL(\hh)=\infty$, for instance $X := \mathbb{N}$, $Y:=\{0,1\}$ and $\hh := \{h: X \to 2^{\{0,1\}}\}$.

Finally, we turn to randomized learners. Here, we have

\begin{proposition}
\label{prp_random}
For all $N \in \mathbb{N}$, $\EM^*_{\hh}(N) \ge \AL(\hh,N)/(\VC(\Lt(\hh))+1)$.
\end{proposition}

(See Appendix~\ref{apx_O1} for the proof.) Together with Theorem~\ref{thm_O1}, this gives us bounds on $\EM^*_{\hh}(N)$ which are tight up to the factor $\VC(\Lt(\hh))+1$. In particular, the same trichotomy of asymptotics in $N$ applies to randomized learners.

\section{Ambiguous Optimal Algorithm}
\label{sec_aoa}
\label{sec:aoa}
We will now describe the algorithm $A_{\text{AOA}}$ which satisfies the upper bound in Theorem~\ref{thm_O1}, i.e., for this algorithm, $\mm^{A_{\text{AOA}}}_{\hh}(N) \le \AL(\hh, N)$.
For this, we will need the following:

\begin{definition}
Consider any function $w: \hh \to \mathbb{N}$ and an ambiguous shattered $\hh$-tree $T=(V,v_0,\Pa,\mathbf{x},\mathbf{y},\mathbf{h},\DE)$.
Then, the \emph{$w$-weighted rank} of $T$ is defined as
\[ \Ra_w(T) := \max_{u \in \Lf} (|\ReSet_T(u)| + w(\mathbf{h}(u))) \]
\end{definition}

The \emph{$w$-weighted ambiguous Littlestone dimension} of $\hh$ (denoted $\AL_w(\hh)$) is defined to be the maximal $w$-weighted rank of an ambiguous shattered $\hh$-tree, or $\infty$ if the rank is unbounded.\footnote{Curiously, \citet{filmusHMM23} also have a notion of Littlestone dimension weighted by a mapping $\hh \to \mathbb{N}$. However, in their setting the weights represent departure from realizability rather than extra mistakes.}
We also denote by $\AL_w(\hh,n)$ the maximal rank of an ambiguous shattered $\hh$-tree of depth at most $n$.
With these ingredients, we can describe the Ambiguous Optimal Algorithm (Algorithm~\ref{alg_aoa}), which is an extension of Littlestone's Standard Optimal Algorithm. The algorithm explicitly depends on the parameter $N$.

On each round, we consider the set of unfalsified hypotheses $\hh_{\text{uf}} \subseteq \hh$. That is, given the history of previous observations $xy \in (X \times Y)^n$, we have
\[ \hh_{\text{uf}} := \{h \in \hh \mid xy \in \cc_h(n)\} \]

We also keep track of the number of mistakes made so far according to each hypothesis. That is, define $w: \hh_{\text{uf}} \to \mathbb{N}$ by
\[ w(h) := |\mm_h^{A_{\text{AOA}}}(xy)| \]

For each possible next label $y' \in Y$, we consider the subset of $\hh_{\text{uf}}$ compatible with it:
\[ \hh_{y'} := \{h \in \hh_{\text{uf}} \mid y' \in h(x_n)\} \]

For each possible prediction $\alpha \in 2^Y$, we consider the updated weight function $w_{\alpha y'}: \hh_{y'} \to \mathbb{N}$ assuming this prediction:
\[ w_{\alpha y'}(h) := w(h) + \max(\mathbf{1}_{\alpha \not\subseteq h(x_n)}, \mathbf{1}_{y' \notin \alpha}) \]

Here and everywhere, $\mathbf{1}_\phi$ stands for $1$ when $\phi$ is true and $0$ when $\phi$ is false. We then choose a prediction by minimaxing the weighted ambiguous Littlestone dimension of the remaining hypotheses on the next step. That is:
\[ A_{\text{AOA}}(N; xy, x_n) := \argmin_{\alpha \in 2^Y} \max_{y' \in Y} \AL_{w_{\alpha y'}}(\hh_{y'}, N-n-1) \]

\begin{algorithm}[t]
\caption{Ambiguous Optimal Algorithm (AOA)}\label{alg_aoa}
\SetKwInOut{KwIn}{Input}
\SetKwInOut{KwOut}{Output}
\KwIn{$N \in \mathbb{N}$, $xy \in (X \times Y)^*$, $x_{\text{new}} \in X$}
\KwOut{Prediction $\alpha^*$}
\BlankLine
Initialize $w: \hh \to \mathbb{N}$ to $\mathbf{0}$\;
$\hh_{\text{uf}} \leftarrow \hh$\;
$n \leftarrow \text{length}(xy)$\;
\For{$k=0$ \KwTo $n$}{
    $D \leftarrow \infty$\;
    $x \leftarrow$ (\uIf{$k<n$}{$x_k$} \Else{$x_{\text{new}}$})\;
    \For{$\alpha \in 2^Y$}{
        $D_\alpha \leftarrow 0$\;
        \For{$y \in Y$}{
            $\hh_y \leftarrow \{h \in \hh_{\text{uf}} \mid y \in h(x)\}$\;
            $w_{\alpha y} \leftarrow w$\;
            \For{$h \in \hh_y$ \KwSty{s.t.} $y \notin \alpha$ \textbf{or} $\alpha \not\subseteq h(x)$}{
                $w_{\alpha y}(h) \leftarrow w_{\alpha y}(h) + 1$\;
            }
            $D_\alpha \leftarrow \max(D_\alpha, \AL_{w_{\alpha y}}(\hh_y, N-k-1))$\;
        }
        \If{$D_\alpha < D$}{
            $\alpha^* \leftarrow \alpha$\;
            $D \leftarrow D_\alpha$\;
        }
    }
    \If{$k<n$}{
        $\hh_{\text{uf}} \leftarrow \{h \in \hh_{\text{uf}} \mid y_k \in h(x_k)\}$\;
        $w \leftarrow w_{\alpha^* y_k}$\;
    }
}
\KwRet{$\alpha^*$}\;
\end{algorithm}

\section{Weighted Aggregation Algorithm}
\label{sec_waa}
\label{sec:waa}
We start from describing an algorithm for \emph{finite} hypothesis classes (i.e., assuming $|\hh| < \infty$). This will later be modified to produce the algorithm for Theorem~\ref{thm_upper_sqrt}. We call it the Weighted Aggregation Algorithm (Algorithm~\ref{alg_waa}), denoted by $A_{\text{WAA}}$.

The algorithm depends on a parameter $\mu \in (1,\infty)$. We keep track of $\hh_{\text{uf}} \subseteq \hh$ and $w:\hh_{\text{uf}} \to \mathbb{N}$ in the same way as in Section~\ref{sec:aoa} (except that we can ignore overconfidence mistakes since they cancel out). Upon receiving instance $x_n \in X$, we compute the vector $q \in \mathbb{R}^{\Lt(\hh)}$ given by
\[ q_\alpha := \sum_{h \in \hh_{\text{uf}}:h(x_n)=\alpha} \mu^{w(h)} \]
We then normalize $q$ in $l_1$ norm to yield $\hat{q} := q/\|q\|_1$. The algorithm's prediction is given by
\begin{equation}
\label{eq_waa}
A_{\text{WAA}}(\mu; xy, x_n) := \left\{y \in Y \;\middle|\; \sum_{\beta \in \Lt(\hh):y \in \beta} \hat{q}_\beta \ge \frac{1+\PD(\Lt(\hh)) \cdot (\mu-1)}{\mu+\PD(\Lt(\hh)) \cdot (\mu-1)}\right\}
\end{equation}

\begin{algorithm}[t]
\caption{Weighted Aggregation Algorithm (WAA)}\label{alg_waa}
\SetKwInOut{KwIn}{Input}
\SetKwInOut{KwOut}{Output}
\KwIn{$\mu \in (1,\infty)$, $xy \in (X \times Y)^*$, $x_{\text{new}} \in X$}
\KwOut{Prediction $\alpha$}
\BlankLine
Initialize $w: \hh \to \mathbb{N}$ to $\mathbf{0}$\;
$\hh_{\text{uf}} \leftarrow \hh$\;
$n \leftarrow \text{length}(xy)$\;
$\nu \leftarrow (1+\PD(\Lt(\hh)) \cdot (\mu-1)) / (\mu+\PD(\Lt(\hh)) \cdot (\mu-1))$\;
\For{$k=0$ \KwTo $n$}{
    $x \leftarrow$ (\uIf{$k<n$}{$x_k$} \Else{$x_{\text{new}}$})\;
    Initialize $q \in \mathbb{R}^{\Lt(\hh)}$ to $\mathbf{0}$\;
    \For{$h \in \hh_{\text{uf}}$}{
        $q_{h(x)} \leftarrow q_{h(x)} + \mu^{w(h)}$\;
    }
    $q \leftarrow q / \|q\|_1$\;
    $\alpha \leftarrow \{y \in Y \mid \sum_{\beta \in \Lt(\hh):y \in \beta} q_\beta \ge \nu\}$\;
    \If{$k<n$}{
        $\hh_{\text{uf}} \leftarrow \{h \in \hh_{\text{uf}} \mid y_k \in h(x_k)\}$\;
        \For{$h \in \hh_{\text{uf}}$ \KwSty{s.t.} $\alpha \not\subseteq h(x)$}{
            $w(h) \leftarrow w(h) + 1$\;
        }
    }
}
\KwRet{$\alpha$}\;
\end{algorithm}

WAA satisfies the following bound.

\begin{theorem}
\label{thm_upper_fin}
Consider any $\mu>1$. Then,
\begin{equation}
\label{eq_thm_upper_fin}
\mm^{A_{\text{WAA}}(\mu)}_{\hh}(N) \le \frac{\ln|\hh| + \frac{1}{2}(\PD(\hh)+1)^2(\mu-1)^2N}{\ln \mu}
\end{equation}
In particular, there exists some global constant $C > 0$ s.t.
\begin{equation}
\label{eq_thm_upper_fin_O}
\inf_\mu \mm^{A_{\text{WAA}}(\mu)}_{\hh}(N) \le C \cdot \PD(\hh)\sqrt{N \log |\hh|}
\end{equation}
\end{theorem}

(See Appendix~\ref{apx_waa} for the proof.) Here, $A_{\text{WAA}}(\mu)$ stands for a WAA learner with the parameter $\mu$.

Now suppose that $\hh$ might be infinite, but $\CL(\hh) < \infty$. Fix the time horizon $N \in \mathbb{N}$. We then construct a new hypothesis class $\hh^{(N)}$ with domain $X^{(N)} := X^{<N} \times X$. Here, the notation $\Sigma^{<N}$ (resp. $\Sigma^{\le N}$) means words over alphabet $\Sigma$ of length less than $N$ (resp. less than or equal to $N$).

Denote $\YP := \Lt(\hh) \setminus \{\emptyset\}$. Let $A_{\text{SOA}}:(X \times \YP)^* \times X \to \YP$ stand for the Standard Optimal Algorithm for hypothesis class $\hh$, regarded as a class of partial functions from $X$ to $\YP$ (see Appendix~\ref{apx_partial} for the definition of SOA for partial function classes).

Consider any $S \subseteq [N]$ with $|S| \le \CL(\hh)$ and $f:S \to \YP$. Define $h_f:X^{(N)} \to \Lt(\hh)$ and $h^*_f:X^{\le N} \to (X \times \YP)^*$ by simultaneous recursion via
\[
\begin{cases}
h_f(\bar{x}, x) &:= \begin{cases} A_{\text{SOA}}(h^*_f(\bar{x}), x) & \text{if } |\bar{x}| \notin S, \\ f(|\bar{x}|) & \text{if } |\bar{x}| \in S \end{cases} \\
h^*_f(\lambda_X) &:= \lambda_{X \times \YP}, \\
h^*_f(\bar{x} x) &:= h^*_f(\bar{x}) x h_f(\bar{x}, x)
\end{cases}
\]
Here, $\bar{x}$ denotes some element of $X^{<N}$, $x$ denotes some element of $X$, $|\bar{x}|$ stands for the length of $\bar{x}$ and $\lambda_\Sigma$ denotes the empty word over alphabet $\Sigma$.

In other words, we imagine running the SOA on some sequence of instances, while using its own predictions as past labels, except for rounds in $S$, in which the labels are determined by $f$.

We now define
\[ \hh^{(N)} := \{h_f \mid S \subseteq [N] \text{ with } |S| \le \CL(\hh) \text{ and } f:S \to \YP\} \]
The hypothesis class $\hh^{(N)}$ can ``imitate" the original class $\hh$ in the following sense.

\begin{lemma}
\label{prp_cover}
For any $x \in X^N$ and $h \in \hh$, if for all $n<N$ it holds that $h(x_n) \neq \emptyset$, then there exists $h' \in \hh^{(N)}$ s.t. for all $n<N$, $h'(x_{:n+1}) = h(x_n)$.
\end{lemma}

(See Appendix~\ref{apx_red} for the proof.) Now consider the learner $A'_{\text{WAA}}$ which is the WAA associated with $\hh^{(N)}$. Using Theorem~\ref{thm_upper_fin} and Lemma~\ref{prp_cover}, it is straightforward to see that $A'_{\text{WAA}}$ satisfies the upper bound of Theorem~\ref{thm_upper_sqrt} w.r.t the hypothesis class $\hh$.

\section{Apple Tasting and Additional Lower Bounds}
\label{sec_apple}
\label{sec:apple}
There is a close connection between ambiguous online learning and so-called \emph{apple tasting} (see \cite{hemboldLL2000}). In apple tasting, there is a hypothesis class $\hh \subseteq \{X \to \{0,1\}\}$ and the learner's output is also in $\{0,1\}$. As in classical online learning, whenever the learner's prediction differs from the true label, this counts as a mistake. In contrast to classical online learning, when the learner's prediction is $0$, the learner receives \emph{no feedback} (whereas if the learner's prediction is $1$, the true label is revealed). We will use the notation $A:(X \times \{0,1\})^* \times X \to \{0,1\}$ for apple tasting learners, where ``no feedback" is represented by $1$.

Given an apple tasting hypothesis $h:X \to \{0,1\}$, we can construct the ambiguous online learning hypothesis $h^{\text{am}}:X \to 2^{\{0,1\}}$ defined by
\[ h^{\text{am}}(x) := \{h(x)\} \cup \{1\} \]
Given an apple tasting hypothesis class $\hh$, we define the ambiguous online learning hypothesis class $\hh^{\text{am}} \subseteq \{X \to 2^{\{0,1\}}\}$ by
\[ \hh^{\text{am}} := \{h^{\text{am}} \mid h \in \hh\} \]
Given an ambiguous learner $A:(X \times Y)^* \times X \to 2^Y$, we define the apple tasting learner $A^{\text{ap}}:(X \times \{0,1\})^* \times X \to \{0,1\}$ by
\[ A^{\text{ap}}(\overline{xy}, x) := \mathbf{1}_{0 \notin A(\overline{xy}, x)} \]
We use the same notations for apple tasting mistake bounds\footnote{Assuming deterministic learners.} as in Section~\ref{sec_setting}, only with $\apm$ instead of $\mm$. (See Appendix~\ref{apx_apple} for the formal definitions.)

We then have,

\begin{proposition}
\label{prp_aap}
For all $h:X \to \{0,1\}$ and $N \in \mathbb{N}$, $\apm^{A^{\text{ap}}}_h(N) \le \mm^A_{h^{\text{am}}}(N)$.
\end{proposition}

(See Appendix~\ref{apx_apple} for the proof.) As obvious corollaries, for any $\hh \subseteq \{X \to \{0,1\}\}$, it holds that $\apm^{A^{\text{ap}}}_{\hh}(N) \le \mm^A_{\hh^{\text{am}}}(N)$ and $\apm^*_{\hh}(N) \le \mm^*_{\hh^{\text{am}}}(N)$.

On the other hand, there's no simple way to turn an apple tasting learner for $\hh$ into an ambiguous online learner for $\hh^{\text{am}}$ with the same mistake bound. However, we have

\begin{proposition}
\label{prp_ap_lower}
For all $N \in \mathbb{N}$, $\apm^*_{\hh}(N) \ge \AL(\hh^{\text{am}}, N)$.
\end{proposition}

Together with Theorem~\ref{thm_O1} and Proposition~\ref{prp_aap}, this implies

\begin{corollary}
\label{crl_apple}
For all $N \in \mathbb{N}$, $\apm^*_{\hh}(N) = \mm^*_{\hh^{\text{am}}}(N)$.
\end{corollary}

One application of this is investigating how tight the bound in Theorem~\ref{thm_upper_fin} is. For any $k,l \ge 2$, denote
\begin{align*}
\mm^*_{k,l}(N) &:= \max_{\hh \subseteq \{X \to 2^{[l]}\} : |\hh|=k} \mm^*_{\hh}(N) \\
\apm^*_k(N) &:= \max_{\hh \subseteq \{X \to \{0,1\}\} : |\hh|=k} \apm^*_{\hh}(N)
\end{align*}
Obviously these functions are non-decreasing in $k,l,N$. By Theorem~\ref{thm_upper_fin}, we have $\mm^*_{k,l}(N) = O(l \sqrt{N \log k})$. By Theorem~\ref{thm_O1} and Proposition~\ref{prp_al_card}, $\mm^*_{k,l}(N) \le k-1$. Also, it's obvious that $\mm^*_{k,l}(N) \le N$. Putting these observations together, we get
\begin{equation}
\label{eq_mkln}
\mm^*_{k,l}(N) = O(\min(N,k,l \sqrt{N \log k}))
\end{equation}
Also, Corollary~\ref{crl_apple} implies that $\mm^*_{k,2}(N) \ge \apm^*_k(N)$. In \cite{hemboldLL2000} we find the following result (called there Corollary 16).

\begin{theorem}[Helmbold-Littlestone-Long]
For all $k \ge 2$ and $N \ge 1$:
\[ \apm_k^*(N) \ge \min\left(\frac{N}{2}, \frac{k-1}{2\sqrt{2}}, \frac{1}{8} \sqrt{\frac{N \ln k}{\ln(1+N/\ln k)}}\right) \]
\end{theorem}

This implies that the bound \eqref{eq_mkln} is tight up to a factor of $O(l \sqrt{\log N})$.

Is it possible to avoid the $l$ factor? We have,

\begin{proposition}
\label{prp_many_labels}
For all $l \ge 1$ and $k \ge 2l$,
\[ \mm_{k,l+1}(\lceil k^2/l \rceil) \ge \frac{k-1}{2} \]
\end{proposition}

(See Appendix~\ref{apx_many_labels} for the proof.) Therefore, it's at best possible to replace $l$ by $\sqrt{l}$.

\section{Conclusion}
We defined a new generalization of the online learning framework: ambiguous online learning (AOL). The theory of apple tasting turned out to also be essentially a special case of AOL. Similarly to how mistake bounds in classical online learning are controlled by Littlestone's ``shattered trees", mistakes bounds in AOL are controlled by ``ambiguous shattered trees". Ambiguous shattered trees are a richer combinatorial object: they need not to be binary or perfect. The upper bound on mistakes via ambiguous shattered trees comes from the AOA, which is broadly similar to the classical SOA. However, additional complexity is involved: the AOA requires us to keep track of \emph{weighted} trees. This is because in classical theory all unfalsified hypotheses are ``on the same footing", but in AOL we have different number of mistakes w.r.t. different unfalsified hypotheses.

Qualitatively, instead of the two possible behaviors of classical theory ($O(1)$ mistakes and $N$ mistakes), we have three possible behaviors, with the new option of $O(\sqrt{N})$ mistakes. The upper bound of this new option is attained using the WAA, an algorithm reminiscent of weighted majority. The coefficient depends both on the classical Littlestone dimension $\CL(\hh)$ and the combinatorial invariant $\PD(\hh)$. As in classical theory, there is at most a constant (w.r.t. $N$) gap between the mistake bounds of deterministic and randomized learners. Our current best characterization of this gap is in terms of $\VC(\Lt(\hh))$. There are many natural questions for further study:

\begin{itemize}
    \item Studying nonrealizable generalizations of AOL. Preliminary attempts by the authors suggest that classical methods do not apply here.
    \item Finding a tight(er) characterization of the coefficient in the $O(\sqrt{N})$ case, or at least closing the $\sqrt{l}$ gap in equation \ref{eq_mkln}.
    \item Closing the $\sqrt{\log N}$ gap between Theorem \ref{thm_upper_sqrt} and Corollary \ref{crl_lower_sqrt}.
    \item We established a trichotomy of mistake bounds for \emph{finite} $Y$, but it's also possible to consider infinite $Y$.
    \item Instead of online learning, we can consider the ambiguous version of offline learning (i.e. VC theory).
    \item Studying stochastic variants of the framework, where each hypothesis predicts a set of probability distributions over labels.
\end{itemize}

\section*{Acknowledgments}

This work was supported by the Machine Intelligence Research Institute in Berkeley, California, the Effective Ventures Foundation USA in San Francisco, California, the Advanced Research + Invention Agency (ARIA) of the United Kingdom, and the Survival and Flourishing Corporation. The author wishes to thank Shay Moran for insightful conversations, especially the suggestions to look into apple tasting and to mimic reduction to finite classes in nonrealizable online learning. She also wishes to thank Bogdan Chornomaz for pointing out the relation between VC dimension and $\Lt$-complexity (Lemma~\ref{lem_vc}). Additionally, she thanks Alexander Appel and Marcus Ogren for reading drafts, pointing out errors, and providing useful suggestions.

\bibliographystyle{plainnat}
\bibliography{main}

\appendix
\section{Properties of Combinatorial Invariants}
\label{apx_com_props}

In this section, we prove the propositions of Section~\ref{sec_inv}.

Proposition~\ref{prp_uniform_rank} is proved by ``trimming" all the unnecessary edges that increase the rank beyond what is required.

\begin{proof}[Proof of Proposition~\ref{prp_uniform_rank}]
Define $r:V \to \mathbb{N}$ by
\[ r(v) := \max_{u \in \Lf: v \in \Pa^*(u)} |\ReSet_T(u) \cap \Pa^*(v)| \]
Denote $r^* := \Ra(T)$. Define $V'$ by
\[ V' := \{v \in V \mid \forall w \in \Pa^*(v) \setminus \{v\}: r(w) < r^*\} \]
Let $\NR' := V' \setminus \{v_0\}$, $\Lf' := V \setminus \Pa(\NR')$. Notice that for any $v \in V'$:
\begin{itemize}
    \item If $r(v) < r^*$, then $\Pa^{-1}(v) \subseteq V'$. It's impossible that $v \in \Lf$ since that would imply $\Ra(T) \le r(v) < r^*$. Hence $\Pa^{-1}(v) \neq \emptyset$ and $v \notin \Lf'$. Also, $|\DE \cap \Pa^{-1}(v) \cap V'| = 1$.
    \item If $r(v) = r^*$, then $\Pa^{-1}(v) \cap V' = \emptyset$ and hence $v \in \Lf'$.
    \item It's impossible that $r(v) > r^*$ since that would imply $r(\Pa(v)) \ge r^*$.
\end{itemize}
Define also $\mathbf{h}':\Lf' \to \hh$ by
\[ \mathbf{h}'(u') := \mathbf{h}\left(\argmax_{u \in \Lf: u' \in \Pa^*(u)} |\ReSet_T(u) \cap \Pa^*(u')|\right) \]
For any $u' \in \Lf'$, we have $r(u')=r^*$, implying that $|\ReSet_{T'}(u')|=r^*$. Therefore, $\Ra(T')=r^*$.
\end{proof}

The notion of a ``subtree" is fairly standard, and will be useful both here and in Appendix~\ref{apx_compact}.

\begin{definition}
Given a rooted tree $T=(V,v_0,\Pa)$, a \emph{subtree} is a rooted tree $T':=(V',v'_0,\Pa')$ s.t.
\begin{itemize}
    \item $V' \subseteq V$
    \item For any $v \in V'$, ${\Pa'}^{*}(v) = V' \cap \Pa^{*}(v)$.
    \item $\Lf' \subseteq \Lf$
\end{itemize}
It is easy to see that $T'$ is uniquely defined\footnote{But not any subset of $V$ corresponds to a subtree: some subsets produce a forest instead, and for some subsets there are leaves which are not leaves of the original tree.} by $V'$. We use the notation $T|_{V'} := T'$. Given any $q \in V$ there is a unique subtree $T^q=(V^q,q,\Pa^q)$ of $T$ s.t. for any $v \in V$, we have $v \in V^q$ if and only if $q \in \Pa^*(v)$.

Given an ambiguous shattered $\hh$-tree $T \Tree$, an \emph{$\hh$-subtree} is an ambiguous shattered tree $T'=(V',v'_0,\Pa',\mathbf{x}',\mathbf{y}',\mathbf{h}',\DE')$ s.t.:
\begin{itemize}
    \item $(V',v'_0,\Pa')$ is a subtree of $(V,v_0,\Pa)$.
    \item $\mathbf{x}' = \mathbf{x}|_{\NL'}$
    \item For any $a \in \NR'$, consider the unique $b \in \Pa^*(a)$ s.t. $\Pa(b)=\Pa'(a)$. Then, $\mathbf{y}'_a = \mathbf{y}_b$. Moreover, $a \in \DE'$ if and only if $b \in \DE$.
    \item $\mathbf{h}' = \mathbf{h}|_{\Lf'}$
\end{itemize}
Again, $T'$ is uniquely defined by $V'$ and we denote $T|_{V'} := T'$. Given any $q \in V$, there is a unique $\hh$-subtree $T^q$ of $T$ s.t. $V^q$ has the same meaning as for unlabeled trees.
\end{definition}

The following definition and lemma are a useful way to simplify ambiguous shattered trees for certain purposes. We use them here and in Appendix~\ref{apx_apple}.

\begin{definition}
Consider an ambiguous shattered $\hh$-tree $T=(V,v_0,\Pa,\mathbf{x},\mathbf{y},\mathbf{h},\DE)$. An edge $a \in \DE$ is called \emph{redundant} when there is no $u \in \Lf$ s.t. $a \in \ReSet_T(u)$. $T$ is called \emph{frugal} when it has no redundant edge.
\end{definition}

\begin{lemma}
\label{lem_frugal}
For every $n \in \mathbb{N}$ and $r \le \AL(\hh,n)$, there exists a frugal $\hh$-tree $T$ with $\Ra(T) \ge r$ and $\Dp(T) \le n$.
\end{lemma}

\begin{proof}
Consider any $\hh$-tree $T=(V,v_0,\Pa,\mathbf{x},\mathbf{y},\mathbf{h},\DE)$. Suppose that $T$ has some redundant $a \in \DE$. Let $v=\Pa(a)$. Denote $V' := (V \setminus V^v) \cup V^a$. Then, $T' := T|_{V'}$ is an $\hh$-subtree of $T$. It's easy to see that $\Ra(T') \ge \Ra(T)$, $\Dp(T') \le \Dp(T)$ and it has at least one less redundant edge. Repeating this procedure eventually produces a frugal tree of rank at least $\Ra(T)$ and depth at most $\Dp(T)$.
\end{proof}

To prove Proposition~\ref{prp_al_card}, we traverse a path going down from the root. At each vertex $v$ on the path, there will be a leaf label $h$ for which the $\DE$-edge of $v$ is relevant, and we'll continue the path through one of the \emph{other} edges which possesses an edge label outside of $h(\mathbf{x}_v)$. The leaf labels we encounter in this way are all different.

\begin{proof}[Proof of Proposition~\ref{prp_al_card}]
Consider any $\hh$-tree $T=(V,v_0,\Pa,\mathbf{x},\mathbf{y},\mathbf{h},\DE)$. Denote $r := \Ra(T)$. We need to show that $r \le |\hh|-1$. By Lemma~\ref{lem_frugal}, we can assume that $T$ is frugal without loss of generality.

If $r \ge 1$, let $a_0 \in \DE$ be s.t. $\Pa(a_0)=v_0$. Since $T$ is frugal, there exists $u_0 \in \Lf$ s.t. $a_0 \in \ReSet_T(u_0)$. Denote $h_0 := \mathbf{h}_{u_0}$ and $x_0 := \mathbf{x}_{v_0}$. It follows that there is $v_1 \in \Pa^{-1}(v_0)$ s.t. $\mathbf{y}_{v_1} \notin h_0(x_0)$. If $r \ge 2$, then $v_1 \notin \Lf$ and we can repeat the same construction starting with $v_1$ instead of $v_0$. Continuing in this manner, we get $(v_0, v_1, \dots, v_r)$, $(x_0, x_1, \dots, x_{r-1})$ and $(h_0, h_1, \dots, h_{r-1})$. Moreover, $v_r \in \Lf$ and we denote $h_r := \mathbf{h}_{v_r}$.

For any $i<r$, denote $y_i := \mathbf{y}_{v_{i+1}}$. For any $j$ with $i<j \le r$, we have $y_i \notin h_i(x_i)$ but $y_i \in h_j(x_i)$, and hence $h_i \neq h_j$. Therefore $|\hh| \ge r+1$.
\end{proof}

The significance of $\VC(\Lt)$ in our context comes from the following lemma.

\begin{lemma}
\label{lem_vc}
For any $Y$-lattice $\Lt$, we have

$$\VC(\Lt)=\max_{A \subseteq Y} \PC_{\Lt}(A)$$
\end{lemma}

\begin{proof}
Consider any $\Lt$-shattered set $A \subseteq Y$. Then, for any $B\subseteq A$, there is some $C \in \Lt$ s.t. $B = A \cap C$. Therefore, $B^{(\Lt)} \cap A = B$. It follows that $\PC_{\Lt}(A)=|A|$. Therefore,

$$\VC(\Lt)\leq\max_{A \subseteq Y} \PC_{\Lt}(A)$$

Conversely, consider any $A \subseteq Y$ and let $k:=\PC_{\Lt}(A)$. Choose $B\subseteq A$ s.t. $|B|=k$ and $A\subseteq B^{(\Lt)}$. Consider any $y \in B$ and denote $B_y := B\setminus\{y\}$. Then, $y \not\in B_y^{(\Lt)}$: otherwise we would have $B\subseteq B_y^{(\Lt)}$ and therefore $A\subseteq B_y^{(\Lt)}$, implying $\PC_{\Lt}(A)\leq k-1$. Consider any $C\subseteq B$ and define

$$\hat{C}:=\bigcap_{y \in B\setminus C} B_y^{(\Lt)}$$

We get that $\hat{C}\in\Lt$ and $C=B\cap \hat{C}$. Since this works for any $C$, we conclude that $B$ is a $\Lt$-shattered set and hence $\VC(\Lt)\geq k$. Therefore,

$$\VC(\Lt)\geq\max_{A \subseteq Y} \PC_{\Lt}(A)$$\end{proof}

It is now trivial to prove Proposition~\ref{prp_upd_card}.

\begin{proof}[Proof of Proposition~\ref{prp_upd_card}]
By Lemma~\ref{lem_vc},

\begin{align*}
\VC(\Lt)&=\max_{A \subseteq Y} \PC_{\Lt}(A)\\
&\geq \max_{A \in \Lt} \PC_{\Lt}(A)\\
&=\PD(\Lt)
\end{align*}
\end{proof}

In order to calculate $\PD(\Lt)$ in Example~\ref{ex_box}, we observe that every $A \in \Lt$ is a hyperrectangle and the $\Lt$-hull of its two extreme corners equals $A$. In order to calculate $\VC(\Lt)$, for any $A \subseteq [n]^d$ we take, for each dimension, a point in $A$ with minimal coordinate along this dimension and a point in $A$ with maximal coordinate. These are $2d$ points whose $\Lt$-hull is $A$.

\begin{proof}[Proof of Example~\ref{ex_box}]
Consider any $a,b \in Y$ s.t. $\forall i<d: a_i \le b_i$. Let $A_{ab} \in \Lt_{d,n}$ be given by
\[ A_{ab} := \{y \in Y \mid \forall i<d: a_i \le y_i \le b_i\} \]
Let $B_{ab} := \{a,b\}$. Then, $B_{ab}^{(\Lt_{d,n})} = A_{ab}$. Hence, $\PC_{\Lt_{d,n}}(A_{ab}) \le 2$ and $\PD(\Lt_{d,n}) \le 2$.

For any $B \subseteq Y$ s.t. $|B|=1$, we have $B \in \Lt_{d,n}$ and hence $B^{(\Lt)}=B$. Therefore, for $a \neq b$, $\PC_{\Lt_{d,n}}(A_{ab}) = 2$ and hence $\PD(\Lt_{d,n})=2$.

For any $A \subseteq Y$, consider
\[ B := \{\argmax_{a \in A} a_i \mid i<d\} \cup \{\argmin_{a \in A} a_i \mid i<d\} \]
Then, $A \subseteq B^{(\Lt_{d,n})}$ and $|B| \le 2d$. Hence, $\PC_{\Lt_{d,n}}(A) \le 2d$ and by Lemma~\ref{lem_vc}, $\VC(\Lt_{d,n}) \le 2d$.

Define $A^* \subseteq Y$ by
\[ A^* := \{a \in Y \mid \exists i<d: a_i \in \{0,2\} \text{ and } \forall j<d: j \neq i \Rightarrow a_j=1\} \]
Then, for any $A \subseteq A^*$, $A^{(\Lt_{d,n})} \cap A^*=A$. Hence, $\PC_{\Lt_{d,n}}((A^*)^{(\Lt_{d,n})}) \ge |A^*| = 2d$ and by Lemma~\ref{lem_vc}, $\VC(\Lt_{d,n}) = 2d$.
\end{proof}

To show Proposition~\ref{prp_arity_upd}, we ``trim" an ambiguous shattered tree to reduce its arity. The key observation is that the definition of relevant edges is s.t. at any vertex $v$, it's enough to retain the edges whose labels have a $\Lt$-hull containing all the original edge labels at $v$, plus the special $\DE$ edge.

\begin{proof}[Proof of Proposition~\ref{prp_arity_upd}]
Let $T_0 \Tree$ be an $\hh$-tree s.t. $\Ra(T_0) \ge r$ and $\Dp(T_0) \le n$. Suppose that $v_1 \in V$ is of arity $m > \VC(\Lt(\hh))+1$. That is, $|\Pa^{-1}(v_1)|=m$. Define
\[ A := \{\mathbf{y}_a \mid a \in \Pa^{-1}(v_1)\} \]
By Lemma~\ref{lem_vc}, there exists $B \subseteq A$ s.t. $A \subseteq B^{(\Lt(\hh))}$ and $|B| \le \VC(\Lt(\hh))$. Define $F \subseteq \NR$ by
\[ F := \{a \in \Pa^{-1}(v_1) \mid \mathbf{y}_a \in B\} \]
Let $a_1 \in \DE$ be the unique edge s.t. $v_1=\Pa(a_1)$, and define $F_1 := F \cup \{a_1\}$. We now construct the $\hh$-tree $T_1=(V' \subseteq V,v_0,\Pa|_{V'},\mathbf{x}|_{\NL'},\mathbf{y}|_{\NR'},\mathbf{h}|_{\Lf'},\DE \cap \NR')$. Here, $V'$ is defined by
\[ V' := \{v \in V \mid F_1 \cap \Pa^*(v) \neq \emptyset \text{ or } v_1 \notin \Pa^*(v) \text{ or } v=v_1\} \]
Consider some $u \in \Lf'$ and denote $\alpha := \mathbf{h}_u(\mathbf{x}_{v_1})$. Consider also some $b \in \Pa'^*(u)$ s.t. $b \notin \ReSet_{T_1}(u)$. If $\Pa'(b) \neq v_1$ then clearly $b \notin \ReSet_{T_0}(u)$. If $\Pa'(b)=v_1$ then for all $a \in F_1 \setminus \{a_1\}$ we have $\mathbf{y}_a \in \alpha$, implying that $B \subseteq \alpha$. It follows that $B^{(\Lt(\hh))} \subseteq \alpha$ and hence $A \subseteq \alpha$. Therefore, $b \notin \ReSet_{T_0}(u)$ in this case as well. We conclude that $\ReSet_{T_0}(u) \subseteq \ReSet_{T_1}(u)$ (in fact they are equal).

Therefore, $\Ra(T_1) \ge \Ra(T_0)$. Obviously, also $\Dp(T_1) \le \Dp(T_0)$. At the same time, the arity of $v_1$ in $T_1$ is $|F_1| \le \VC(\Lt(\hh))+1$. This construction can be repeated until we get the desired $T$.
\end{proof}

In a classical shattered tree, the same instance cannot appear multiple times along a single path from the root to a leaf. For ambiguous shattered trees this is not so. However, the following lemma establishes that, in some sense, there is a limit to how many times you can usefully repeat the same instance in a single path.

\begin{lemma}
\label{lem_repeating_x}
Let $T \Tree$ be a frugal $\hh$-tree. Consider any $v^* \in \NL$, and denote $x^* := \mathbf{x}_{v^*}$. Define $A \subseteq Y$ by
\[ A := \{\mathbf{y}_b \mid b \in \Pa^*(v^*) \setminus \{v_0\} \text{ and } \mathbf{x}_{\Pa(b)}=x^*\} \]
Then, there exists $a \in \Pa^{-1}(v^*)$ s.t. $\mathbf{y}_a \notin A^{(\Lt(\hh))}$.
\end{lemma}

\begin{proof}
Let $a^* \in \DE$ be s.t. $\Pa(a^*)=v^*$. Since $a^*$ is non-redundant, there exists $u \in \Lf$ and $a \in \NR$ s.t. $a^* \in \Pa^*(u)$, $\Pa(a)=v^*$ and $\mathbf{y}_a \notin \mathbf{h}_u(x^*)$. Denote $y^* := \mathbf{y}_a$ and $B := \mathbf{h}_u(x^*)$. For any $b \in \Pa^*(v^*) \setminus \{v_0\}$ s.t. $\mathbf{x}_{\Pa(b)}=x^*$, we have $b \in \Pa^*(u)$ and therefore $\mathbf{y}_b \in B$. We conclude that $A \subseteq B$. Since $B \in \Lt(\hh)$, it follows that $A^{(\Lt(\hh))} \subseteq B$. On the other hand $y^* \notin B$, implying that $y^* \notin A^{(\Lt(\hh))}$.
\end{proof}

Using Lemma~\ref{lem_repeating_x}, we can construct a chain in $\Lt(\hh)$ out of the edge labels associated with repeated instances within a single path. Hence, the number of such repeated instances is bounded by the length of $\Lt(\hh)$. This allows us to prove Proposition~\ref{prp_al_dom}.

\begin{proof}[Proof of Proposition~\ref{prp_al_dom}]
Consider any $\hh$-tree $T \Tree$. We need to show that $\Ra(T) \le l|X|$. By Lemma~\ref{lem_frugal}, we can assume w.l.o.g. that $T$ is frugal. Construct the sequence $\{v_k \in V\}_{k \le n}$ as follows. We start with $v_0$ and proceed by recursion. Define $A_k \subseteq Y$ by
\[ A_k := \{\mathbf{y}_a \mid a \in \Pa^*(v_k) \setminus \{v_0\} \text{ and } \mathbf{x}_{\Pa(a)} = \mathbf{x}_{v_k}\} \]
Choose $v_{k+1} \in V$ to be s.t. $\Pa(v_{k+1})=v_k$ and $\mathbf{y}_{v_{k+1}} \notin A_k^{(\Lt(\hh))}$. By Lemma~\ref{lem_repeating_x}, this is possible as long as $v_k \notin \Lf$ (hence $v_n \in \Lf$).

For every $x \in X$, let $I_x \subseteq [n]$ be defined by
\[ I_x := \{k<n \mid \mathbf{x}_{v_k}=x\} \]
For any $k<n$, denote $B_k := A_k^{(\Lt(\hh))}$. Notice that whenever $i,j \in I_x$ and $i<j$, we have $\mathbf{y}_{v_{i+1}} \notin B_i$, $\mathbf{y}_{v_{i+1}} \in A_j$ and therefore $B_j \setminus B_i \neq \emptyset$. Hence, $B_i \subset B_j$.

Let $k_x := \max I_x$ and define $A_{(x)} := A_{k_x} \cup \{\mathbf{y}_{v_{k_x+1}}\}$ and $B_{(x)} := A_{(x)}^{(\Lt(\hh))}$. Then, $\mathbf{y}_{v_{k_x+1}} \notin B_{k_x}$, $\mathbf{y}_{v_{k_x+1}} \in A_{(x)}$ and hence $B_{k_x} \subset B_{(x)}$. We constructed a chain of length $|I_x|+1$ and hence $|I_x| \le l-1$. Since $[n] = \bigcup_x I_x$, we get $n \le l|X|$. Therefore $|\ReSet_T(v_n)| \le n \le l|X|$ and $\Ra(T) \le l|X|$.
\end{proof}

In order to prove Proposition~\ref{prp_CL_vs_AL}, we choose the edge labels for $T'$ as elements of the corresponding edge labels of $T$, and select $\DE$, in such a way as to make all edges relevant for all leaves under them.

\begin{proof}[Proof of Proposition~\ref{prp_CL_vs_AL}]
Consider any $v \in \NL$. Let $a,b \in \NR$ be s.t. $\Pa(a)=\Pa(b)=v$ and $a \neq b$. Let $\alpha := \mathbf{y}_a$ and $\beta := \mathbf{y}_b$. We know that $\alpha \neq \beta$. Assume w.l.o.g. that $\alpha \setminus \beta \neq \emptyset$. Then, we set $b \in \DE$ (and $a \notin \DE$). Choose any $y \in \alpha \setminus \beta$ and set $\mathbf{y}'_a := y$. Choose any $z \in \beta$ and set $\mathbf{y}'_b := z$.

These choices guarantee that for any $u \in \Lf$ s.t. $b \in \Pa^*(u)$, we have $b \in \ReSet_{T'}(u)$. Indeed, denote $x := \mathbf{x}_v$ and $h := \mathbf{h}_u$. Then $h(x)=\beta$ and therefore $\mathbf{y}'_a \notin h(x)$. Moreover, for any $w \in \Lf$ s.t. $a \in \Pa^*(w)$, we have $a \in \ReSet_{T'}(w)$: simply because $a \notin \DE$.

Repeating these steps for every $v \in \NL$, we fully specify $T'$. For every $u \in \Lf$, we have $\ReSet_{T'}(u) = \Pa^*(u) \setminus \{v_0\}$ and hence $|\ReSet_{T'}(u)|=\Dp(u)$. It follows that $\Ra(T')=\Dp(T)$.
\end{proof}

\section{Littlestone Dimension for Partial Function Classes}
\label{apx_partial}

In this section, we briefly review the theory of online learning for partial function classes from \cite{alonHHM2021}.

Let $X$ be the domain and $Y$ the label set. A \emph{partial function class} is some $\hh \subseteq \{X \to Y \cup \{\bot\}\}$. Here ``$\bot$" is a special symbol that means ``undefined". The intended meaning of $h(x)=\bot$ is, the instance $x$ \emph{cannot} appear (with any label).

Mistakes are defined the same way as in classical (realizable) online learning, except that the adversary is not allowed to select instance $x \in X$ if $h^* \in \hh$ is the true hypothesis and $h^*(x)=\bot$. As in classical online learning, the optimal mistake bound is described by (the appropriate generalization of) Littlestone dimension.

\begin{definition}
A \emph{shattered $\hh$-tree} is a tuple $T:=(V,v_0,\Pa,\mathbf{x},\mathbf{y},\mathbf{h})$, where $(V,v_0,\Pa)$ is a rooted tree and $\mathbf{x}:\NL \to X$, $\mathbf{y}:\NR \to Y$, $\mathbf{h}:\Lf \to \hh$ are labelings. We require that $T$ is binary, i.e. for any $v \in \NL$, $|\Pa^{-1}(v)|=2$. We also require that $T$ is perfect, i.e. for any $u,w \in \Lf$, $|\Pa^*(u)|=|\Pa^*(w)|$. Moreover, the labelings have to satisfy the following conditions:
\begin{itemize}
    \item For any $a,b \in \NR$, if $a \neq b$ but $\Pa(a)=\Pa(b)$ then $\mathbf{y}_a \neq \mathbf{y}_b$.
    \item For any $u \in \Lf$ and $a \in \Pa^*(u) \setminus \{v_0\}$: $\mathbf{y}_a = \mathbf{h}_u(\mathbf{x}_{\Pa(a)})$.
\end{itemize}
\end{definition}

The \emph{Littlestone dimension of $\hh$} (denoted $\Li(\hh)$) is defined to be the maximal depth of a shattered $\hh$-tree, or $\infty$ if the depth is unbounded.

We define the \emph{Standard Optimal Algorithm} $A_{\text{SOA}}:(X \times Y)^* \times X \to Y$ as follows. Consider any $n \in \mathbb{N}$, $xy \in (X \times Y)^n$ and $x^* \in X$. For each $y \in Y$, define $\hh_y$ by
\[ \hh_y := \{h \in \hh \mid h(x^*)=y \text{ and } \forall k<n: h(x_k)=y_k\} \]
Then, we set
\[ A_{\text{SOA}}(xy,x^*) := \argmax_{y \in Y} \Li(\hh_y) \]
Analogously to the classical setting, we have

\begin{theorem}
\label{thm_soa}
For any partial function class $\hh$ and $N \in \mathbb{N}$, the following mistake bound holds.
\[ \mm^{A_{\text{SOA}}}_{\hh}(N) \le \Li(\hh) \]
\end{theorem}

Here, we use the same notation for mistake bounds as in our own ambiguous setting. The abuse of notation is not great, since $\hh$ can be interpreted as an ambiguous class where every $h \in \hh$ is replaced by $h':X \to 2^Y$ defined by
\[ h'(x) := \begin{cases} \emptyset & \text{if } h(x)=\bot, \\ \{h(x)\} & \text{if } h(x) \neq \bot \end{cases} \]
Theorem~\ref{thm_soa} is essentially Theorem 47 from \cite{alonHHM2021}, except that we don't assume $|Y|=2$. Extending the proof is trivial (and standard for total function classes, see Theorem 24 in \cite{danielySBS15}) and we omit it.

\section{Mistakes and the Ambiguous Littlestone Dimension}
\label{apx_O1}

In this section we prove Theorem~\ref{thm_O1} and Proposition~\ref{prp_random}.

We start from showing the lower bound. This is accomplished by interpreting every ambiguous shattered tree $T$ as a strategy by which an adversary can force $\Ra(T)$ mistakes within $\Dp(T)$ rounds. Specifically, this strategy works by traversing the tree from the root to one of the leaves, and presenting the learner with the instances and labels associated with the vertices and edges, respectively, along the path. When the learner's prediction rules out the label of an edge $a$ going down from the current vertex $v$, we choose $a$ to continue the path, forcing an overconfidence mistake. In other cases, we choose the $\DE$ edge $b$, forcing an underconfidence mistake whenever $b$ is relevant for the leaf $u$ we ultimately reach. The resulting trace is compatible with the hypothesis labeling $u$.

\begin{lemma}
\label{lem_O1_lower}
Consider any ambiguous shattered $\hh$-tree $T$. Let $n:=\Dp(T)$ and $r:=\Ra(T)$. Then, $\mm^*_{\hh}(n) \ge r$.
\end{lemma}

\begin{proof}
Consider any learner $A:(X \times Y)^* \times X \to 2^Y$. Construct $xy \in (X \times Y)^*$ and $v \in V^*$ recursively as follows. We start from $v_0$. Suppose we constructed $xy_{:k}$ and $v_{:(k+1)}$ for some $k \in \mathbb{N}$. If $v_k \in \Lf$, the process ends (i.e. $|xy|=k$ and $|v|=k+1$). Otherwise, set $x_k:=\mathbf{x}_{v_k}$ and consider $\alpha_k:=A(xy_{:k},x_k)$. If there is some $a \in \Pa^{-1}(v_k)$ s.t. $\mathbf{y}_a \notin \alpha_k$, set $y_k:=\mathbf{y}_a$ and $v_{k+1}:=a$. Otherwise, let $b$ be the unique element of $\DE \cap \Pa^{-1}(v_k)$. Set $y_k:=\mathbf{y}_b$ and $v_{k+1}:=b$.

Denote $m:=|xy|$. Let $h:=\mathbf{h}_{v_m}$. Then, $xy \in \PC_h(m)$. Moreover, for any $k<m$, if $v_{k+1} \in \ReSet_T(v_m)$ then $k \in \mm^A_h(xy)$. Indeed, either $y_k \notin \alpha_k$, which is an overconfidence mistake, or $v_{k+1} \in \DE$ and there is some $a \in \Pa^{-1}(\Pa(v_{k+1}))$ s.t. $\mathbf{y}_a \notin h(x_k)$ (because $v_{k+1}$ is relevant to $v_m$) and $\mathbf{y}_a \in \alpha_k$ (by construction of $v_{k+1}$), making it an underconfidence mistake.

We conclude that $\mm_{\hh}^A(n) \ge \mm_h^A(n) \ge \mm_h^A(m) \ge |\ReSet_T(v_m)| \ge r$. Since this holds for any $A$, we get $\mm_{\hh}^*(n) \ge r$.
\end{proof}

As an obvious consequence, we have $\mm^*_{\hh}(n) \ge \AL(\hh,n)$.

The proof of Proposition~\ref{prp_random} is very similar to the proof of Lemma~\ref{lem_O1_lower}. The difference is, since the learner is randomized, we can no longer choose edges that \emph{guarantee} to produce mistakes whenever they are relevant. Instead, we choose edges which come with a satisfactory lower bound on the \emph{probability} of a mistake. Given that we've already seen that for any prediction $\alpha \subseteq Y$, there is always at least one edge that produces a mistake (when it's relevant), we can get a lower bound of inverse arity. On the other hand, we can upper bound the arity using Proposition~\ref{prp_arity_upd}.

\begin{proof}[Proof of Proposition~\ref{prp_random}]
Consider any $N \in \mathbb{N}$ and $r \le \AL(\hh,N)$. By Proposition~\ref{prp_arity_upd}, there exists an ambiguous shattered $\hh$-tree $T$ with $\Dp(T) \le N$, $\Ra(T) \ge r$ and $\Ar(T) \le \VC(\Lt(\hh))+1$. Denote $q:=\Ar(T)$.

Consider any randomized learner $R:(X \times Y)^* \times X \to \Delta 2^Y$. Construct $xy \in (X \times Y)^*$ and $v \in V^*$ recursively as follows. We start from $v_0$. Suppose we constructed $xy_{:k}$ and $v_{:(k+1)}$ for some $k \in \mathbb{N}$. If $v_k \in \Lf$, the process ends (i.e. $|xy|=k$ and $|v|=k+1$). Otherwise, set $x_k:=\mathbf{x}_{v_k}$ and consider $\theta_k:=R(xy_{:k},x_k)$. Define
\[ \beta_k := \{\mathbf{y}_b \mid b \in \Pa^{-1}(v_k)\} \]
Define $F_k:2^Y \to 2^{\NR}$ by
\[ F_k(\alpha) := \{b \in \Pa^{-1}(v_k) \mid \mathbf{y}_b \notin \alpha \text{ or } (b \in \DE \text{ and } \beta_k \subseteq \alpha)\} \]
Notice that $F_k(\alpha)$ is always non-empty: either $\beta_k \setminus \alpha \neq \emptyset$ and then there is some $b \in \Pa^{-1}(v_k)$ s.t. $\mathbf{y}_b \notin \alpha$ and hence $b \in F_k(\alpha)$, or $\beta_k \subseteq \alpha$ and then $\DE \cap \Pa^{-1}(v_k) \subseteq F_k(\alpha)$. Define
\[ v_{k+1} := \argmax_{b \in \Pa^{-1}(v_k)} \Pr_{\alpha \sim \theta_k}[b \in F_k(\alpha)] \]
Notice that the maximal probability on the right hand side is at least $1/q$, since $F_k(\alpha) \neq \emptyset$ and $|\Pa^{-1}(v_k)| \le q$. Set $y_k:=\mathbf{y}_{v_{k+1}}$.

Denote $m:=|xy|$ and $u:=v_m$. Let $h:=\mathbf{h}_u$. Then, $xy \in \PC_h(m)$. Consider any $k<m$ s.t. $v_{k+1} \in \ReSet_T(u)$. The probability that $v_{k+1} \in F_k(\alpha)$ for $\alpha \sim \theta_k$ is at least $1/q$. Also, $v_{k+1} \in F_k(\alpha)$ implies that $R$ made a mistake. We conclude that $\EM_{\hh}^R(N) \ge \EM_h^R(N) \ge \EM_h^R(m) \ge \frac{1}{q} |\ReSet_T(v_m)| \ge r/q$. Since it holds for any $R$, we get $\EM_{\hh}^*(N) \ge r/q$.
\end{proof}

In the following, we will use the notations
\begin{align*}
\hh_{\text{uf}}^{\overline{xy}} &:= \{h \in \hh \mid \overline{xy} \in \PC_h(k)\} \\
w^{\overline{xy}}_A(h) &:= |\mm_h^A(\overline{xy})|
\end{align*}
Here, $A$ is some learner and $w^{\overline{xy}}$ is a function from $\hh_{\text{uf}}^{\overline{xy}}$ to $\mathbb{N}$.

The following lemma establishes that the weighted (by number of prior mistakes) ambiguous Littlestone dimension of the set of unfalsified hypotheses cannot increase when the learner follows the AOA. This is proved by showing that for each round, we can construct an ambiguous shattered tree by joining at a new root several ambiguous shattered trees associated with the next round for different possible predictions.

\begin{lemma}
\label{lem_aoa_inv}
Fix $N \in \mathbb{N}$. Consider some $k<N$, $\overline{xy} \in \PC_{\hh}(k)$, $x^* \in X$ and $y^* \in Y$ s.t. $\overline{xy}x^*y^* \in \PC_{\hh}(k+1)$. We claim that
\begin{equation}
\label{eq_lem_O1_inv}
\AL_{w_{A_{\text{AOA}}(N)}^{\overline{xy}x^*y^*}}(\hh_{\text{uf}}^{\overline{xy}x^*y^*}, N-(k+1)) \le \AL_{w_{A_{\text{AOA}}(N)}^{\overline{xy}}}(\hh_{\text{uf}}^{\overline{xy}}, N-k)
\end{equation}
\end{lemma}

For the rest of Appendix~\ref{apx_O1}, we denote $A^*:=A_{\text{AOA}}(N)$ and omit the subscript $A^*$ under $w$.

\begin{proof}[Proof of Lemma~\ref{lem_aoa_inv}]
For any $\alpha \in 2^Y$ and $y \in Y$, define $\tilde{w}^{\alpha y}:\hh_{\text{uf}}^{\overline{xy}x^*y} \to \mathbb{N}$ by
\[ \tilde{w}^{\alpha y}(h) := w^{\overline{xy}}(h) + \max(\mathbf{1}_{y \notin \alpha}, \mathbf{1}_{\alpha \not\subseteq h(x^*)}) \]
Define $r \in \mathbb{N}$ by
\[ r := \min_{\alpha \in 2^Y} \max_{y \in Y} \AL_{\tilde{w}^{\alpha y}}(\hh_{\text{uf}}^{\overline{xy}x^*y}, N-(k+1)) \]
$A^*(\overline{xy}, x^*)$ is defined to achieve the minimum above, hence if $y^*$ is the observed label,
\begin{equation}
\label{eq_lem_O1_upper_lhs}
\AL_{w^{\overline{xy}x^*y^*}}(\hh_{\text{uf}}^{\overline{xy}x^*y^*}, N-(k+1)) \le r
\end{equation}
For any $\alpha \in 2^Y$, there must be some $y^\alpha \in Y$ s.t.
\[ \AL_{\tilde{w}^{\alpha y^\alpha}}(\hh_{\text{uf}}^{\overline{xy}x^*y^\alpha}, N-(k+1)) \ge r \]
Let $T^\alpha=(V^\alpha,v^\alpha_0,\Pa^\alpha,\mathbf{x}^\alpha,\mathbf{y}^\alpha,\mathbf{h}^\alpha,\DE^\alpha)$ be an $\hh_{\text{uf}}^{\overline{xy}x^*y^\alpha}$-tree s.t. $\Dp(T^\alpha) \le N-(k+1)$ and $\Ra_{\tilde{w}^{\alpha y^\alpha}}(T^\alpha) \ge r$. Define
\begin{equation}
\label{eq_hat_alpha}
\hat{\alpha} := \{y \in Y \mid \exists \alpha \in 2^Y: y=y^\alpha \text{ and } y \notin \alpha\}
\end{equation}
Denote $\hat{y} := y^{\hat{\alpha}}$. Notice that the definition of $\hat{\alpha}$ implies that $\hat{y} \in \hat{\alpha}$. Indeed, assume to the contrary that $\hat{y} \notin \hat{\alpha}$. Then, $\hat{y}$ satisfies the condition on $y$ on the right hand side of equation \eqref{eq_hat_alpha} with $\alpha=\hat{\alpha}$ ($\hat{y}=y^{\hat{\alpha}}$ by definition of $\hat{y}$, and $\hat{y} \notin \hat{\alpha}$ by assumption), implying that $\hat{y} \in \hat{\alpha}$, a contradiction.

For each $y \in \hat{\alpha}$, let $\alpha_y \in 2^Y$ be s.t. $y=y^{\alpha_y}$ and $y \notin \alpha_y$. We will now construct a new tree by joining some of the $T^\alpha$ at a new root. Denote $\alpha' := \hat{\alpha} \setminus \{\hat{y}\}$. The vertices are
\[ V := \{v_0\} \cup V^{\hat{\alpha}} \cup \bigcup_{y \in \alpha'} V^{\alpha_y} \]
The parent mapping $\Pa$ is defined s.t. for any $\beta \in \{\hat{\alpha}\} \cup \{\alpha_y\}_{y \in \alpha'}$ and $v \in V^\beta \setminus \{v_0^\beta\}$, we have $\Pa(v) := \Pa^\beta(v)$ and $\Pa(v_0^\beta) := v_0$. The labels for the $T^\beta$ subtrees remain as they are. In addition, we set $\mathbf{x}_{v_0} := x^*$, $\mathbf{y}_{v_0^{\hat{\alpha}}} := \hat{y}$ and for any $y \in \alpha'$, $\mathbf{y}_{v_0^{\alpha_y}} := y$. Finally, we set
\[ \DE := \{v_0^{\hat{\alpha}}\} \cup \DE^{\hat{\alpha}} \cup \bigcup_{y \in \alpha'} \DE^{\alpha_y} \]
Consider any $u \in \Lf^{\hat{\alpha}}$. Then, $\mathbf{h}_u \in \hh_{\text{uf}}^{\overline{xy}x^*\hat{y}}$ and hence $\mathbf{y}_{v_0^{\hat{\alpha}}} = \hat{y} \in \mathbf{h}_u(x^*)$, as it should. Similarly, consider any $y \in \alpha'$ and $u \in \Lf^{\alpha_y}$. Then, $\mathbf{h}_u \in \hh_{\text{uf}}^{\overline{xy}x^*y}$ and hence $\mathbf{y}_{v_0^{\alpha_y}} = y \in \mathbf{h}_u(x^*)$. Therefore, $T$ is an $\hh_{\text{uf}}^{\overline{xy}}$-tree.

Consider again some $u \in \Lf^{\hat{\alpha}}$ and denote $h:=\mathbf{h}_u$. Since $\hat{y} \notin \hat{\alpha}$, we have $\tilde{w}^{\hat{\alpha}\hat{y}}(h) = w^{\overline{xy}}(h) + \mathbf{1}_{\hat{\alpha} \not\subseteq h(x^*)}$. Moreover, $v_0^{\hat{\alpha}} \in \DE$ and therefore $\hat{\alpha} \not\subseteq h(x^*)$ iff $v_0^{\hat{\alpha}} \in \ReSet_T(u)$. Hence
\[ |\ReSet_T(u)| + w^{\overline{xy}}(h) = |\ReSet_{T^{\hat{\alpha}}}(u)| - 1 + \mathbf{1}_{v_0^{\hat{\alpha}} \in \ReSet_T(u)} + w^{\overline{xy}}(h) = |\ReSet_{T^{\hat{\alpha}}}(u)| + \tilde{w}^{\hat{\alpha}\hat{y}}(h) \ge r \]
Now, consider any $y \in \alpha'$ and $u \in \Lf^{\alpha_y}$, and denote $h := \mathbf{h}_u$. Since $y \notin \alpha_y$, we have $\tilde{w}^{\alpha_y y}(h) = w^{\overline{xy}}(h)+1$. Moreover, $v_0^{\alpha_y} \notin \DE$ and therefore $v_0^{\alpha_y} \in \ReSet_T(u)$. Hence
\[ |\ReSet_T(u)| + w^{\overline{xy}}(h) = |\ReSet_{T^{\alpha_y}}(u)| + 1 + w^{\overline{xy}}(h) = |\ReSet_{T^{\alpha_y}}(u)| + \tilde{w}^{\alpha_y y}(h) \ge r \]
We conclude that $\Ra_{w^{\overline{xy}}}(T) \ge r$. Also, $\Dp(T) \le N-(k+1)+1=N-k$. It follows that
\begin{equation}
\label{eq_lem_O1_upper_rhs}
\AL_{w^{\overline{xy}}}(\hh_{\text{uf}}^{\overline{xy}}, N-k) \ge r
\end{equation}
Putting inequalities \eqref{eq_lem_O1_upper_lhs} and \eqref{eq_lem_O1_upper_rhs} together, we get inequality \eqref{eq_lem_O1_inv}.
\end{proof}

Using the invariant from Lemma~\ref{lem_aoa_inv}, it is straightforward to prove an upper bound for the AOA.

\begin{lemma}
\label{lem_O1_upper}
For all $N \in \mathbb{N}$, $\mm^{A^*}_{\hh}(N) \le \AL(\hh,N)$.
\end{lemma}

\begin{proof}
Chaining Lemma~\ref{lem_aoa_inv} over $k<N$, we get that for any $\overline{xy} \in \PC_{\hh}(N)$
\begin{equation}
\label{eq_lem_O1_upper_chain}
\AL_{w^{\overline{xy}}}(\hh_{\text{uf}}^{\overline{xy}},0) \le \AL(\hh,N)
\end{equation}
Now, consider any $h \in \hh$ s.t. $\overline{xy} \in \PC_h(N)$. We have the $\hh$-tree $T_h:=(\{v_0\},v_0,\Pa_\emptyset,\mathbf{x}_\emptyset,\mathbf{y}_\emptyset,\mathbf{h}_h,\emptyset)$ where $\mathbf{h}_h(v_0):=h$ and mappings $\Pa_\emptyset,\mathbf{x}_\emptyset,\mathbf{y}_\emptyset$ have empty domains. Observing that $\Ra_{w^{\overline{xy}}}(T_h) = w^{\overline{xy}}(h) = |\mm^{A^*}_h(\overline{xy})|$, we conclude
\[ \AL_{w^{\overline{xy}}}(\hh_{\text{uf}}^{\overline{xy}},0) \ge |\mm^{A^*}_h(\overline{xy})| \]
By inequality \eqref{eq_lem_O1_upper_chain}, we get
\[ |\mm^{A^*}_h(\overline{xy})| \le \AL(\hh,N) \]
The result follows.
\end{proof}

Putting Lemma~\ref{lem_O1_lower} and Lemma~\ref{lem_O1_upper} together, we immediately get Theorem~\ref{thm_O1}.

\section{Trimming Shattered Trees}
\label{apx_compact}

In this section, we prove Theorem~\ref{thm_compact}. To achieve this, it's convenient to consider the following, slightly stronger, statement.

\begin{lemma}
\label{lem_compact}
Let $T \Tree$ be an $\hh$-tree, $N \in \mathbb{N}$ and assume that $\Ra(T) \ge N^2$. Then, there exists $T'$, an $\hh$-subtree of $T$, s.t. $\Ra(T') \ge N$ and $\Dp(T') \le N^2$.
\end{lemma}

The idea of the proof is judiciously selecting $N$ edges out of the $\DE$-path starting at the root, and then using induction to trim the subtrees attached to these edges.

\begin{proof}[Proof of Lemma~\ref{lem_compact}]
We proceed by induction on $N$.

For $N=0$, we take any $u \in \Lf$ and set $T':=T^u$. Obviously, $\Dp(T')=0$.

Now, assume the claim holds for some $N$. Consider an $\hh$-tree $T \Tree$ s.t. $\Ra(T) \ge (N+1)^2$. Construct the sequence $\{v_i \in V\}_{i \le d}$, for some $d \in \mathbb{N}$, as follows:
\begin{itemize}
    \item $v_0$ is the root.
    \item For any $i<d$, we recursively define $v_{i+1}$ to be the unique vertex s.t. $v_{i+1} \in \DE$ and $v_i = \Pa(v_{i+1})$.
    \item The recursion terminates at $v_d \in \Lf$.
\end{itemize}
Let $i \le d$ be minimal s.t. there exists some $u^* \in \Lf$ for which $v_i \in \Pa^*(u^*)$ and
\begin{equation}
\label{eq_lem_compact}
|\{j \le i \mid v_j \in \ReSet_T(u^*)\}| = N+1
\end{equation}
This $i$ necessarily exists, since $|\ReSet_T(v_d)| \ge (N+1)^2 \ge N+1$.

Fix some $u^* \in \Lf$ s.t. the identity \eqref{eq_lem_compact} holds. Consider any $j<i$ s.t. $v_{j+1} \in \ReSet_T(u^*)$. Consider also some $w \in \Pa^{-1}(v_j) \setminus \{v_{j+1}\}$. Let's examine the $\hh$-subtree $T^w$. For any $u \in \Lf^w$, the minimality of $i$ implies
\[ |\ReSet_T(u) \cap \{v_k\}_{k \le d}| \le N \]
Moreover
\[ \ReSet_T(u) = (\ReSet_T(u) \cap \{v_k\}_{k \le d}) \cup \{w\} \cup \ReSet_{T^w}(u) \]
On the other hand, $\Ra(T) \ge (N+1)^2$ and hence $|\ReSet_T(u)| \ge (N+1)^2$. It follows that $|\ReSet_{T^w}(u)| \ge (N+1)^2-N-1 \ge N^2$. Since this holds for any $u \in \Lf^w$, we conclude that $\Ra(T^w) \ge N^2$. Hence, we can apply the induction hypothesis to $T^w$, yielding $\tilde{T}^w$, an $\hh$-subtree of $T^w$ with $\Ra(\tilde{T}^w) \ge N$ and $\Dp(\tilde{T}^w) \le N^2$.

Define $S \subseteq V$ by
\[ S := \{v_j \mid j<i \text{ and } v_{j+1} \in \ReSet_T(u^*)\} \]
We can now construct the desired $T'$. The set of vertices is defined to be
\[ V' := S \cup \{u^*\} \cup \bigcup_{w \in \Pa^{-1}(S) \setminus \DE} \tilde{V}^w \]
Here, $\tilde{V}^w$ is the set of vertices of $\tilde{T}^w$. We set $T' := T|_{V'}$.

It's easy to see that the leaves of $T'$ are
\[ \Lf' = \{u^*\} \cup \bigcup_{w \in \Pa^{-1}(S) \setminus \DE} \tilde{\Lf}^w \]
Let's establish the rank of $T'$. Clearly, $\Pa'(\ReSet_{T'}(u^*)) = S$. The identity \eqref{eq_lem_compact} implies that $|S|=N+1$ and hence $|\ReSet_{T'}(u^*)|=N+1$. Moreover, for any $w \in \Pa^{-1}(S) \setminus \DE$ and $u \in \tilde{\Lf}^w$, we have $\tilde{v}_0^w \in \ReSet_{T'}(u)$, since $w \notin \DE$ and therefore $\tilde{v}_0^w \notin \DE'$. (Here, $\tilde{v}_0^w$ is the root of $\tilde{T}^w$.) Also, $\tilde{v}_0^w \notin \ReSet_{\tilde{T}^w}(u)$, since $\tilde{v}_0^w \notin \tilde{\NR}^w$. Therefore,
\[ \ReSet_{\tilde{T}^w}(u) \cup \{\tilde{v}_0^w\} \subseteq \ReSet_{T'}(u) \]
Hence, $|\ReSet_{T'}(u)| \ge N+1$ and we conclude that $\Ra(T') \ge N+1$.

Finally, let's establish the depth of $T'$. Clearly, ${\Pa'}^{*}(u^*) = S \cup \{u^*\}$, and hence the depth of $u^*$ is $N+1 \le (N+1)^2$. For any $w \in \Pa^{-1}(S) \setminus \DE$ and $u \in \tilde{\Lf}^w$, we have
\[ {\Pa'}^{*}(u) \subseteq \tilde{\Pa}^{*w}(u) \cup S \]
Hence, $|{\Pa'}^{*}(u)| \le \Dp(\tilde{T}^w) + |S| \le N^2+N+1 \le (N+1)^2$. Therefore, $\Dp(T') \le (N+1)^2$.
\end{proof}

Lemma~\ref{lem_compact} immediately implies Theorem~\ref{thm_compact}.

\section{Mistake Bound for WAA}
\label{apx_waa}

In this section we prove Theorem~\ref{thm_upper_fin}.

Fix $\mu \in (1,\infty)$ and $\nu \in [1/\mu, 1)$. Let $A^*$ be the learner which generalizes the WAA, where we replace equation \eqref{eq_waa} with the more general rule
\[ A^*(xy,x_n) := \left\{y \in Y \middle| \sum_{\beta \in \Lt(\hh):y \in \beta} \hat{q}_\beta \ge \nu\right\} \]
We will omit the subscript $A^*$ under $w$. Also, we will use the notations
\begin{align*}
Z(\overline{xy}) &:= \sum_{h \in \hh_{\text{uf}}^{\overline{xy}}} \mu^{w^{\overline{xy}}(h)} \\
\hat{q}^{\overline{xy}x}_\alpha &:= \frac{1}{Z(\overline{xy})} \sum_{h \in \hh_{\text{uf}}^{\overline{xy}}:h(x)=\alpha} \mu^{w^{\overline{xy}}(h)}
\end{align*}

The following lemma shows that the rate with which $Z$ grows along an $\hh$-realizable trace can be bounded. Here's the idea of the proof. If the new label $y^*$ is outside the predicted set $\alpha^*$, then a substantial fraction (measured by $\mu^w$) of hypotheses become falsified: if a lot of hypotheses were consistent with $y^*$, the learner would assign $y^*$ into its prediction. If $y^*$ is inside $\alpha^*$, then the fraction of hypotheses that gain a mistake cannot be too great. Because, we can choose a set $\beta \subseteq \alpha^*$ of ``pivots" whose cardinality is at most $\PD(\hh)$, and only hypotheses that exclude at least one of the pivots gain a mistake, but each pivot has to be contained in a substantial fraction of hypotheses since it was selected into $\alpha^*$.

\begin{lemma}
\label{lem_waa_inv}
Consider some $k \in \mathbb{N}$, $\overline{xy} \in \PC_{\hh}(k)$, $x^* \in X$ and $y^* \in Y$ s.t. $\overline{xy}x^*y^* \in \PC_{\hh}(k+1)$. Then,
\begin{equation}
\label{eq_lem_waa_inv}
Z(\overline{xy}x^*y^*) \le \max(\mu\nu, 1+\PD(\hh) \cdot (\mu-1)(1-\nu))Z(\overline{xy})
\end{equation}
\end{lemma}

\begin{proof}
Denote $\alpha^* := A^*(\overline{xy}, x^*)$. There are two cases: $y^* \in \alpha^*$ and $y^* \notin \alpha^*$.

First, consider the case $y^* \notin \alpha^*$. Then,
\begin{align*}
Z(\overline{xy}x^*y^*) 
  &= \sum_{h \in \hh_{\text{uf}}^{\overline{xy}x^*y^*}} \mu^{w^{\overline{xy}x^*y^*}(h)} \\
  &= \sum_{h \in \hh_{\text{uf}}^{\overline{xy}}:y^* \in h(x^*)} \mu^{w^{\overline{xy}}(h)+1} \\
  &= \mu \sum_{\alpha \in \Lt(\hh):y^* \in \alpha} \sum_{h \in \hh_{\text{uf}}^{\overline{xy}}:h(x^*)=\alpha} \mu^{w^{\overline{xy}}(h)} \\
  &= \mu \sum_{\alpha \in \Lt(\hh):y^* \in \alpha} Z(\overline{xy})\hat{q}^{\overline{xy}x^*}_\alpha \\
  &= \mu Z(\overline{xy}) \sum_{\alpha \in \Lt(\hh):y^* \in \alpha} \hat{q}^{\overline{xy}x^*}_\alpha \\
  &< \mu Z(\overline{xy}) \cdot \nu
\end{align*}

Now, consider the case $y^* \in \alpha^*$. Choose $\beta \subseteq \alpha^*$ s.t. $\alpha^* \subseteq \beta^{(\Lt(\hh))}$ and $|\beta| \le \PD(\hh)$. Then, for any $h \in \hh$, if $\beta \subseteq h(x^*)$ then $\alpha^* \subseteq h(x^*)$. Such a hypothesis $h$ doesn't gain a mistake. Conversely, if $\alpha^* \setminus h(x^*) \neq \emptyset$ then $\beta \setminus h(x^*) \neq \emptyset$. Denote
\[ Z_\beta := \sum_{h \in \hh_{\text{uf}}^{\overline{xy}}:\beta \setminus h(x^*) \neq \emptyset} \mu^{w^{\overline{xy}}(h)} \]
We get,
\begin{align*}
Z(\overline{xy}x^*y^*)
  &= \sum_{h \in \hh_{\text{uf}}^{\overline{xy}x^*y^*}} \mu^{w^{\overline{xy}x^*y^*}(h)} \\
  &= \sum_{h \in \hh_{\text{uf}}^{\overline{xy}x^*y^*}:\beta \setminus h(x^*) = \emptyset} \mu^{w^{\overline{xy}x^*y^*}(h)} + \sum_{h \in \hh_{\text{uf}}^{\overline{xy}x^*y^*}:\beta \setminus h(x^*) \neq \emptyset} \mu^{w^{\overline{xy}x^*y^*}(h)} \\
  &= \sum_{h \in \hh_{\text{uf}}^{\overline{xy}x^*y^*}:\beta \setminus h(x^*) = \emptyset} \mu^{w^{\overline{xy}}(h)} + \mu \sum_{h \in \hh_{\text{uf}}^{\overline{xy}x^*y^*}:\beta \setminus h(x^*) \neq \emptyset} \mu^{w^{\overline{xy}}(h)} \\
  &\le \sum_{h \in \hh_{\text{uf}}^{\overline{xy}}:\beta \setminus h(x^*) = \emptyset} \mu^{w^{\overline{xy}}(h)} + \mu \sum_{h \in \hh_{\text{uf}}^{\overline{xy}}:\beta \setminus h(x^*) \neq \emptyset} \mu^{w^{\overline{xy}}(h)} \\
  &= Z(\overline{xy}) - Z_\beta + \mu Z_\beta \\
  &= Z(\overline{xy}) + (\mu-1)Z_\beta
\end{align*}
Moreover,
\begin{align*}
Z_\beta
  &= \sum_{\alpha \in \Lt(\hh):\beta \setminus \alpha \neq \emptyset} \sum_{h \in \hh_{\text{uf}}^{\overline{xy}}:h(x^*)=\alpha} \mu^{w^{\overline{xy}}(h)} \\
  &= \sum_{\alpha \in \Lt(\hh):\beta \setminus \alpha \neq \emptyset} Z(\overline{xy}) \hat{q}^{\overline{xy}x^*}_\alpha \\
  &\le \sum_{y \in \beta} \sum_{\alpha \in \Lt(\hh):y \notin \alpha} Z(\overline{xy}) \hat{q}^{\overline{xy}x^*}_\alpha \\
  &= Z(\overline{xy}) \sum_{y \in \beta} \sum_{\alpha \in \Lt(\hh):y \notin \alpha} \hat{q}^{\overline{xy}x^*}_\alpha \\
  &= Z(\overline{xy}) \sum_{y \in \beta} \left(1 - \sum_{\alpha \in \Lt(\hh):y \in \alpha} \hat{q}^{\overline{xy}x^*}_\alpha\right) \\
  &\le Z(\overline{xy}) \sum_{y \in \beta} (1-\nu) \\
  &\le (1-\nu) \PD(\hh) Z(\overline{xy})
\end{align*}
Combining the two inequalities, we conclude,
\begin{align*}
Z(\overline{xy}x^*y^*)
  &\le Z(\overline{xy}) + (\mu-1)Z_\beta \\
  &\le Z(\overline{xy}) + (\mu-1)(1-\nu) \PD(\hh) Z(\overline{xy}) \\
  &= (1+\PD(\hh) \cdot (\mu-1)(1-\nu))Z(\overline{xy})
\end{align*}
\end{proof}

We are now ready to complete the proof of Theorem~\ref{thm_upper_fin}. In order to show inequality \eqref{eq_thm_upper_fin}, we apply Lemma~\ref{lem_waa_inv} on each of the $N$ rounds, for the value of $\nu$ that corresponds to equation \eqref{eq_waa}. Equation \eqref{eq_thm_upper_fin_O} follows from the inequality \eqref{eq_thm_upper_fin} when we choose appropriate values of $\mu$.

\begin{proof}[Proof of Theorem~\ref{thm_upper_fin}]
Denote $D:=\PD(\Lt(\hh))$. For the WAA, we have
\[ \nu = \frac{1+D(\mu-1)}{\mu+D(\mu-1)} \]
Consider the right hand side of inequality \eqref{eq_lem_waa_inv} of Lemma~\ref{lem_waa_inv}. For the first argument of the maximum, we have
\[ \mu\nu = \frac{\mu+D\mu(\mu-1)}{\mu+D(\mu-1)} \]
For the second argument of the maximum, we have
\begin{align*}
1+D(\mu-1)(1-\nu) 
  &= 1+D(\mu-1) \cdot \frac{\mu-1}{\mu+D(\mu-1)} \\
  &= \frac{\mu + D(\mu-1)+D(\mu-1)^2}{\mu+D(\mu-1)} \\
  &= \frac{\mu + D\mu(\mu-1)}{\mu+D(\mu-1)}
\end{align*}
Hence, Lemma~\ref{lem_waa_inv} takes the form
\[ Z(\overline{xy}x^*y^*) \le \frac{\mu + D\mu(\mu-1)}{\mu+D(\mu-1)} \cdot Z(\overline{xy}) \]
Taking a logarithm, we get
\begin{align}
\ln Z(\overline{xy}x^*y^*)
  &\le \ln Z(\overline{xy}) + \ln \mu + \ln(1+D(\mu-1)) - \ln(\mu+D(\mu-1)) \nonumber \\
  &\le \ln Z(\overline{xy}) + \mu-1 + D(\mu-1) - \ln(\mu+D(\mu-1)) \nonumber \\
  &\le \ln Z(\overline{xy}) + \frac{1}{2}(\mu-1+D(\mu-1))^2 \nonumber \\
  &= \ln Z(\overline{xy}) + \frac{1}{2}(D+1)^2(\mu-1)^2 \label{eq_prf_thm_upper_fin}
\end{align}
Here, we used the inequality $x - \ln x - 1 \le \frac{1}{2}(x-1)^2$ for $x \ge 1$. This is seen by observing that both sides vanish with their first derivatives at $x=1$, while for second derivatives we have $1/x^2 \le 1$ when $x \ge 1$.

Repeatedly using inequality \eqref{eq_prf_thm_upper_fin}, we conclude that for any $\overline{xy} \in \PC_{\hh}(N)$
\begin{align*}
\ln Z(\overline{xy}) 
  &\le \ln Z(\lambda) + \frac{1}{2}(D+1)^2(\mu-1)^2 N \\
  &= \ln|\hh| + \frac{1}{2}(D+1)^2(\mu-1)^2 N
\end{align*}
For any $h \in \hh$ s.t. $\overline{xy} \in \PC_h(N)$, we have $h \in \hh_{\text{uf}}^{\overline{xy}}$ and hence $Z(\overline{xy}) \ge \mu^{w^{\overline{xy}}(h)}$. Therefore,
\[ \ln \mu^{w^{\overline{xy}}(h)} \le \ln|\hh| + \frac{1}{2}(D+1)^2(\mu-1)^2 N \]
Rearranging, we get
\[ w^{\overline{xy}}(h) \le \frac{\ln|\hh| + \frac{1}{2}(D+1)^2(\mu-1)^2 N}{\ln \mu} \]
This establishes inequality \eqref{eq_thm_upper_fin}.

In order to get equation \eqref{eq_thm_upper_fin_O}, we consider two cases. The first case is $(D+1)^2 N \ge \ln|\hh|$. Then, we set
\[ \mu := 1 + \frac{1}{D+1} \sqrt{\frac{\ln|\hh|}{N}} \]
In particular, $\mu \le 2$. By the concavity of the logarithm, $\ln \mu \ge (\ln 2)(\mu - 1)$. We get,
\begin{align*}
w^{\overline{xy}}(h)
  &\le \frac{1}{\ln 2} \left( \frac{\ln|\hh|}{\mu-1} + \frac{1}{2}(D+1)^2(\mu-1)N \right) \\
  &= \frac{3}{2 \ln 2}(D+1)\sqrt{N \ln|\hh|}
\end{align*}
Notice that if $D=0$ then $\Lt(\hh)=\{Y\}$, implying that $A^* \equiv Y$ and hence $\mm^{A^*}_{\hh}(N)=0$. So, in any case the right hand side can be bounded by $C D \sqrt{N \log|\hh|}$.

The second case is $(D+1)^2 N < \ln|\hh|$. Then,
\[ (D+1)\sqrt{N \ln|\hh|} \geq (D+1)\sqrt{N \cdot (D+1)^2 N} = (D+1)^2 N \]
The right hand side is obviously an upper bound on $w^{\overline{xy}}(h)$.
\end{proof}

\section{Reduction to Finite Classes}
\label{apx_red}

In this section we prove Theorem~\ref{thm_upper_sqrt}.

We start with Lemma~\ref{prp_cover}. The latter is a fairly straightforward consequence of Theorem~\ref{thm_soa}. This is essentially the same method as the reduction to finite hypothesis classes used when proving the Littlestone dimension regret bound for unrealizable online learning (see e.g. \cite{shalevB2014}, Lemma 21.13).

\begin{proof}[Proof of Lemma~\ref{prp_cover}]
Define $\{S_n \subseteq [n]\}_{n \le N}$ recursively by setting $S_0 := \emptyset$, and for $n < N$
\[ S_{n+1} := \begin{cases} S_n & \text{if } A_{\text{SOA}}(h_{f_n}^*(x_{:n}),x_n)=h(x_n), \\ S_n \cup \{n\} & \text{otherwise} \end{cases} \]
where for each $n \le N$, the function $f_n:S_n \to \YP$ is given by $f_n(k) := h(x_k)$ for $k \in S_n$. We used the assumption $h(x_k) \neq \emptyset$ in the definition of $f_n$.

In other words, we imagine running the SOA on the sequence of instances $x$, while allowing it direct access to the hypothesis $h$ on past instances (a richer feedback than our setting allows). The sets $S$ record the rounds where SOA makes a mistake (i.e. predicts the label set $h(x)$ incorrectly), while $f$ records the corrections its predictions require.

Clearly, $S_N = \mm^{A_{\text{SOA}}}_{h^*_{f_N}}(x)$.\footnote{We don't need a subscript for $\mm$ here because in classical online learning we don't need to know the true/reference hypothesis to determine when mistakes occur.} By Theorem~\ref{thm_soa}, $|S_N| \le \CL(\hh)$. We can therefore take $h' := h_{f_N}$.
\end{proof}

Theorem~\ref{thm_upper_sqrt} now follows from Lemma~\ref{prp_cover} and Theorem~\ref{thm_upper_fin}, via bounding the cardinality of $\hh^{(N)}$.

\begin{proof}[Proof of Theorem~\ref{thm_upper_sqrt}]
Consider any $N \in \mathbb{N}$, $h \in \hh$ and $xy \in \cc_h(N)$. For all $n<N$, $y_n \in h(x_n)$ and in particular $h(x_n) \neq \emptyset$. By Lemma~\ref{prp_cover}, there exists $h' \in \hh^{(N)}$ s.t. for all $n<N$, $h'(x_{:n+1}) = h(x_n)$. Define $\tilde{x} \in (X^{(N)})^N$ by $\tilde{x}_n := x_{:n+1}$. Then, $\tilde{x}y \in \cc_{h'}(N)$. By Theorem~\ref{thm_upper_fin},
\[ |\mm^{A_{\text{WAA}}}_{h'}(\tilde{x}y)| \le C_1 \cdot \PD(\hh)\sqrt{N \log |\hh^{(N)}|} \]
Here, $C_1>0$ is a global constant and $A_{\text{WAA}}$ refers to the WAA learner for the hypothesis class $\hh^{(N)}$.

Moreover, there is a global constant $C_2>1$ s.t.
\[ \log|\hh^{(N)}| \le C_2 \CL(\hh)\log(|\Lt(\hh)| N) \]
Therefore,
\[ |\mm^{A_{\text{WAA}}}_{h'}(\tilde{x}y)| \le C_1 \sqrt{C_2} \cdot \PD(\hh)\sqrt{\CL(\hh)N \log(|\Lt(\hh)| N)} \]
Now, consider the learner $A'_{\text{WAA}}$ defined by $A'_{\text{WAA}}(xy_{:n}, x_n) := A_{\text{WAA}}(\tilde{x}y_{:n}, \tilde{x}_n)$. We have $\mm^{A_{\text{WAA}}}_{h'}(\tilde{x}y) = \mm^{A'_{\text{WAA}}}_{h}(xy)$ and hence $A'_{\text{WAA}}$ satisfies the required mistake bound. Indeed, for any $n<N$, the prediction of $A'_{\text{WAA}}$ for instance $x_n$ is based on the prediction of $A_{\text{WAA}}$ for instance $\tilde{x}_n$, and $h(x_n)=h'(\tilde{x}_n)$, so the corresponding mistake conditions coincide.
\end{proof}

\section{Apple Tasting Details}
\label{apx_apple}

In this section, we prove Proposition~\ref{prp_aap} and Proposition~\ref{prp_ap_lower}.

First, let's formally define the apple tasting setting. Let $X$ be the domain. Given a learner $A: (X \times \{0,1\})^* \times X \to \{0,1\}$, a hypothesis $h:X \to \{0,1\}$, some $n \in \mathbb{N}$ and $xy \in (X \times \{0,1\})^n$, $xy$ is said to be \emph{compatible} with $h$ and $A$ when for all $k < n$, if $A(xy_{:k}, x_k)=0$ then $y_k=1$ and if $A(xy_{:k}, x_k)=1$ then $y_k = h(x_k)$. Here, $y$ represents the feedbacks received by the learner (rather than the true labels). We denote by $\cc_h^A(n)$ the set of all such $xy$.

Given $xy \in \cc_h^A(n)$, we define the set of \emph{mistakes} that $A$ makes on $xy$ relative to $h$ by
\[ \apm_h^A(xy) := \{k<n \mid A(xy_{:k}, x_k) \neq h(x_k)\} \]
We then define
\[ \apm_h^A(N) := \max_{xy \in \cc_h^A(N)} |\apm_h^A(xy)| \]
Given a hypothesis class $\hh \subseteq \{X \to \{0,1\}\}$, we also define
\begin{align*}
\apm_{\hh}^A(N) &:= \max_{h \in \hh} \apm_h^A(N) \\
\apm_{\hh}^*(N) &:= \min_A \apm_{\hh}^A(N)
\end{align*}

We now come to the proof of Proposition~\ref{prp_aap}. Informally, we're comparing the apple tasting learner $A^{\text{ap}}$ interacting with an environment that obeys $h$ to the ambiguous learner $A$ interacting with an environment that obeys $h^{\text{am}}$. We argue that erroneously predicting $0$ in apple tasting corresponds to an underconfidence mistake in ambiguous online learning, whereas erroneously predicting $1$ in apple tasting corresponds to an overconfidence mistake in ambiguous online learning. There are two potentially substantial differences. First, in apple tasting we receive no feedback when we predict $0$, whereas in ambiguous online learning we still receive a label. However, in the latter setting, the adversary can always choose the label $1$, and when the learner predicts $\{0,1\}$ there is no incentive for the adversary to choose $0$. Hence, effectively there is no useful feedback in the ambiguous case as well. Second, in apple tasting we receive full information about $h(x)$ when we predict $1$, whereas in ambiguous online learning the adversary may choose the label $1$ even if $h(x)=\{0,1\}$ (but in this case we incur no mistake). However, this only gives the adversary additional options, so the ambiguous version can only be harder.

\begin{proof}[Proof of Proposition~\ref{prp_aap}]
Consider any $xy \in \cc_h^{A^{\text{ap}}}(N)$. For any $k<N$, there are two cases: $A^{\text{ap}}(xy_{:k}, x_k)=0$ and $A^{\text{ap}}(xy_{:k}, x_k)=1$. If $A^{\text{ap}}(xy_{:k}, x_k)=0$ then $y_k=1$, and in particular $y_k \in h^{\text{am}}(x_k)$, since $1 \in h^{\text{am}}(x)$ for all $x \in X$. If $A^{\text{ap}}(xy_{:k}, x_k)=1$ then $y_k=h(x_k)$, and in particular $y_k \in h^{\text{am}}(x_k)$, since $\{h(x)\} \subseteq h^{\text{am}}(x)$ for all $x \in X$. In either case $y_k \in h^{\text{am}}(x_k)$. Hence, $xy \in \cc_{h^{\text{am}}}(N)$.

Consider any $k \in \apm_h^{A^{\text{ap}}}(xy)$. Again, there are two cases: $A^{\text{ap}}(xy_{:k}, x_k)=0$ and $A^{\text{ap}}(xy_{:k}, x_k)=1$.

If $A^{\text{ap}}(xy_{:k}, x_k)=0$ then $h(x_k)=1$ and $0 \in A(xy_{:k}, x_k)$. It follows that $h^{\text{am}}(x_k)=\{1\}$ and $0 \in A(xy_{:k}, x_k) \setminus h^{\text{am}}(x_k)$. In particular, $A(xy_{:k}, x_k) \not\subseteq h^{\text{am}}(x_k)$ and hence $k \in \mm_{h^{\text{am}}}^A(xy)$.

If $A^{\text{ap}}(xy_{:k}, x_k)=1$ then $h(x_k)=0$ and $0 \notin A(xy_{:k}, x_k)$. Hence, $y_k=0$, because $xy \in \cc_h^{A^{\text{ap}}}(N)$. In particular, $y_k \notin A(xy_{:k}, x_k)$ and hence $k \in \mm_{h^{\text{am}}}^A(xy)$.

In either case $k \in \mm_{h^{\text{am}}}^A(xy)$. Therefore, $\apm_h^{A^{\text{ap}}}(xy) \subseteq \mm_{h^{\text{am}}}^A(xy)$. In particular, $|\apm_h^{A^{\text{ap}}}(xy)| \le |\mm_{h^{\text{am}}}^A(xy)| \le \mm_{h^{\text{am}}}^A(N)$. Since this holds for any $xy \in \cc_h^{A^{\text{ap}}}(N)$, we get $\apm_h^{A^{\text{ap}}}(N) \le \mm_{h^{\text{am}}}^A(N)$.
\end{proof}

Now, let's prove Proposition~\ref{prp_ap_lower}. The idea is that any frugal ambiguous shattered tree can be interpreted as an adversary strategy for the \emph{apple tasting} setting. The strategy works by going down the $0$ edge when the learner predicts $1$ and going down the $1$ edge (which has to be the $\DE$ edge) when the learner predicts $0$. In the former case, we always get a mistake (and the edge is always relevant). In the latter case, we get a mistake if and only if the edge is relevant to the ultimate leaf.

\begin{proof}[Proof of Proposition~\ref{prp_ap_lower}]
Consider any apple tasting learner $A:(X \times \{0,1\})^* \times X \to \{0,1\}$. Let $T \Tree$ be an ambiguous shattered $\hh^{\text{am}}$-tree with $\Dp(T) \le N$ and $\Ra(T)=r$. It is sufficient to show that $\apm^A_{\hh}(N) \ge r$.

By Lemma~\ref{lem_frugal}, we can assume without loss of generality that $T$ is frugal. In particular, for every $v \in \NL$ it holds that $|\Pa^{-1}(v)|>1$. Since there are only 2 labels, this implies $|\Pa^{-1}(v)|=2$. For $y \in \{0,1\}$, we will denote by $C_y(v)$ the unique $a \in \NR$ s.t. $\Pa(a)=v$ and $\mathbf{y}_a=y$. Moreover, since $T$ is frugal, $C_1(v) \in \DE$. Indeed, it's impossible to have $C_0(v) \in \DE$, since for any $u \in \Lf$ s.t. $C_0(v) \in \Pa^*(u)$ we have $\mathbf{h}^{\text{am}}_u(\mathbf{x}_v)=\{0,1\}$ and hence $C_0(v) \in \DE$ would imply $C_0(v) \notin \ReSet_T(u)$.

Construct $xy \in (X \times \{0,1\})^*$ and $v \in V^*$ recursively as follows. We start from $v_0$. Suppose we constructed $xy_{:k}$ and $v_{:(k+1)}$ for some $k \in \mathbb{N}$. If $v_k \in \Lf$, the process ends (i.e. $|xy|=k$ and $|v|=k+1$). Otherwise, set $x_k:=\mathbf{x}_{v_k}$. Let $z_k:=A(xy_{:k},x_k)$. Set $y_k:=1-z_k$ and $v_{k+1}:=C_{y_k}(v_k)$.

Denote $m:=|xy|$. Let $h:=\mathbf{h}_{v_m}$. Then, $xy \in \PC^A_h(m)$. Moreover, for any $k<m$, if $v_{k+1} \in \ReSet_T(v_m)$ then $k \in \apm^A_h(xy)$. Indeed, either $y_k=0$ in which case $z_k=1$ while $h^{\text{am}}(x_k)=\{0,1\}$ and hence $h(x_k)=0$, or $y_k=1$ in which case $z_k=0$ while $h^{\text{am}}(x_k)=\{1\}$ (it cannot be $\{0,1\}$ because $v_{k+1} \in \DE \cap \ReSet_T(v_m)$) and hence $h(x_k)=1$.

We conclude that $|\apm_h^A(xy)| \ge r$ while $|xy| \le N$. Therefore, $\apm_{\hh}^A(N) \ge r$.
\end{proof}

\section{Lower Bound for Many Labels}
\label{apx_many_labels}

In this section, we prove Proposition~\ref{prp_many_labels}.

Since the definition of $\mm_{k,l}^*(N)$ is a worst-case over \emph{all} hypothesis classes of cardinality $k$, we can think of it as a game between the learner and the adversary, in which the adversary is allowed to decide on every round, for each hypothesis $h_i$ with $i<k$, what is the set of labels $h_i(x) \subseteq 2^{[l]}$. The different possible moves of the adversary can be encoded as different instances $x$.

We prove Proposition~\ref{prp_many_labels} by constructing a tree $T$ and defining a hypothesis class $\hh$ based on $T$ s.t. $T$ becomes an ambiguous shattered $\hh$-tree which certifies the lower bound. $T$ can be thought of as representing a particular strategy for the adversary in the above game.

Specifically, the strategy works roughly as follows: on every round, we choose a set $B$ of $l$ hypotheses, out of the set $A$ of unfalsified hypotheses, by picking the hypotheses $j \in A$ with the \emph{least} number of prior mistakes $m(j)$. Fixing a bijection $f$ between $B$ and $[l+1] \setminus \{0\}$, we let every $j \in B$ predict the label set $[l+1] \setminus \{f(j)\}$. Every hypothesis in $A \setminus B$ predicts the full label set $Y=[l+1]$. If the learner predicts $\alpha \subseteq [l+1]$ then a label $y \in [l+1] \setminus \alpha$ is selected, and all surviving hypotheses in $A$ gain 1 mistake, while at most one hypothesis is falsified. If the learner predicts $\alpha=[l+1]$ then the label $y=0$ is selected, nothing is falsified and all hypotheses in $B$ gain a mistake.

Under this strategy, if the learner always predicts $[l+1]$, after approx $k^2/l$ rounds all hypotheses gain approx $k$ mistakes (it takes $k/l$ rounds for $B$ to cycle through all hypotheses). If the learner predicts anything else, it can only make things worse because more hypotheses gain mistakes, whereas to falsify most hypotheses you have to allow the remaining hypothesis to gain approx $k$ mistakes. Details follow.

\begin{proof}[Proof of Proposition~\ref{prp_many_labels}]
Given any finite non-empty set $A$ and $m:A \to \mathbb{N}$, we recursively construct a tree $T^{(m)}=(V^{(m)},v^{(m)}_0,\Pa^{(m)})$. It will be equipped with $\DE^{(m)} \subseteq \NR^{(m)}$ and a labeling $\mathbf{y}^{(m)}:\NR^{(m)} \to [l+1]$ satisfying the usual conditions for ambiguous shattered trees. It will also be equipped with a labeling $\mathbf{h}^{(m)}:\Lf^{(m)} \to A$.

For the base of the recursion, assume that either $\max m \ge k$ or $|A|=1$. Then, set $V^{(m)} := \{v_0^{(m)}\}$ and choose $j^* := \mathbf{h}^m_{v_0^{(m)}}$ to maximize $m(j^*)$.

For the step of the recursion, we can assume that $|A| \ge 2$. Define $B \subseteq A$ as follows. If $|A| \ge l$, choose any $B \subseteq A$ that minimizes $\max m|_B$ subject to $|B|=l$. If $|A|<l$, set $B:=A$.

Define $m^{(0)}:A \to \mathbb{N}$ by
\[ m^{(0)}(j) := \begin{cases} m(j)+1 & \text{if } j \in B, \\ m(j) & \text{if } j \notin B \end{cases} \]
Choose an injection $f:B \to [l+1] \setminus \{0\}$. For each $i \in f(B)$, define $m^{(i)}:A \setminus \{f^{-1}(i)\} \to \mathbb{N}$ by $m^{(i)} := m|_{A \setminus \{f^{-1}(i)\}} + 1$.

Define $T^m$ to consist of the subtrees $T^{(m^{(0)})}$ and $\{T^{(m^{(i)})}\}_{i \in f(B)}$ joined together at a new root. Define $\DE^m$ by
\[ \DE^m := \DE^{(m^{(0)})} \cup \bigcup_{i \in f(B)} \DE^{(m^{(i)})} \cup \{v_0^{(m^{(0)})}\} \]
Set $\mathbf{y}_{v_0^{(m^{(i)})}} := i$ for any $i \in f(B) \cup \{0\}$. The rest of the $\mathbf{y}$ and $\mathbf{h}$ labelings are inherited from the subtrees. This completes the definition of $T^m$.

Define $m^*:[k] \to \mathbb{N}$ by $m^* \equiv 0$ and let $T:=T^{m^*}$. Define $X:=\NL$ and equip $T$ with the labeling $\mathbf{x}:\NL \to X$ given by $\mathbf{x}_v:=v$. For any $j < k$, define $h_j:X \to 2^{[l+1]}$ as follows. Consider any $v \in \NL$. Then, $T^v=T^{m^v}$ for some $A^v \subseteq [k]$ and $m^v:A^v \to \mathbb{N}$. Let $B^v \subseteq A^v$ and $f^v:B^v \to [l+1] \setminus \{0\}$ be as in the definition of $T^{m^v}$. If $j \in B^v$, set $h_j(v) := [l+1] \setminus \{f^v(j)\}$. Otherwise, set $h_j(v):=[l+1]$.

For every $v \in V$, define $\bar{m}^v \in \mathbb{Q}$ by
\[ \bar{m}^v := \frac{1}{|A^v|} \sum_{j \in A^v} m^v(j) \]
Several observations:
\begin{itemize}
    \item For every $v \in V$, there is some $m_0^v \in \mathbb{N}$ s.t. $m^v(A^v) \subseteq \{m_0^v, m_0^v+1\}$ (due to our starting $m^*$ and how $B$ is defined).
    \item For every $v,w \in V$, if $v=\Pa(w)$ then $\bar{m}^w \ge \bar{m}^v + l/k$ and $\bar{m}^w + \frac{1}{2}|A^w| \ge \bar{m}^v + \frac{1}{2}|A^v|$. Indeed, either $|A^w|=|A^v|$ and $\bar{m}^w = \bar{m}^v + |B^v|/|A^v|$, or $|A^w|=|A^v|-1$ and $\bar{m}^w \ge \bar{m}^v - 1/|A^v| + 1 \ge \bar{m}^v + 1/2$.
\end{itemize}

Define $\hh := \{h_j \mid j<k\}$. It's easy to see that $T$ is an ambiguous shattered $\hh$-tree (we abuse notation by identifying $j$ with $h_j$ in the leaf labeling $\mathbf{h}$). Moreover, $\Dp(T) \le \lceil k^2/l \rceil$. No path starting at the root can be longer, because $\bar{m}^v$ increases by at least $l/k$ on every edge, and if it reaches $k$ the path terminates.

Finally, $\Ra(T) \ge (k-1)/2$. Indeed, given $u \in \Lf$ and $a \in \DE \cap \Pa^*(u)$, we have $a \in \ReSet_T(u)$ if and only if $\mathbf{h}_u \in B^{(\Pa(a))}$. By induction, we see that for any $v \in \Pa^*(u)$, $m^v(\mathbf{h}_u) = |\ReSet_T(u) \cap \Pa^*(v)|$. In particular, $|\ReSet_T(u)| = m^u(\mathbf{h}_u)$. Since $u$ is a leaf, either $\max m^u \ge k$ in which case $m^u(\mathbf{h}_u) \ge k$ or $|A^u|=1$, in which case $m^u(\mathbf{h}_u) = \bar{m}^u \ge (k-1)/2$ since $\bar{m}^v + \frac{1}{2}|A^v|$ is non-decreasing down the path.
\end{proof}

\end{document}